\documentclass[manuscript,screen,review=false]{acmart}

\usepackage{amsmath,amsfonts}
\usepackage{algorithmic}
\usepackage{algorithm}
\usepackage{array}
\usepackage[caption=false,font=footnotesize,labelfont=rm,textfont=rm]{subfig}
\usepackage{textcomp}
\usepackage{stfloats}
\usepackage{url}
\usepackage{verbatim}
\usepackage{graphicx}
\usepackage{booktabs}
\usepackage{threeparttable}
\usepackage{makecell}
\usepackage{color, multirow, amsmath, amsthm, mathrsfs}
\usepackage{bbding}
\usepackage{fontawesome}
\usepackage{makecell}
\usepackage{hyperref}
\AtBeginDocument{%
  \providecommand\BibTeX{{%
    \normalfont B\kern-0.5em{\scshape i\kern-0.25em b}\kern-0.8em\TeX}}}

\setcopyright{acmlicensed}
\acmJournal{TOMM}
\acmYear{2024} \acmVolume{1} \acmNumber{1} \acmArticle{1} \acmMonth{1}\acmDOI{10.1145/3640344}

\newcommand{\ie}{\textit{i}.\textit{e}.}
\newcommand{\et}{\textit{e}\textit{t} \textit{a}\textit{l}.}

\begin{document}

\title{Viewport-Unaware Blind Omnidirectional Image Quality Assessment: A Flexible and Effective Paradigm}


\author{Jiebin Yan}
\email{jiebinyan@foxmail.com}
\orcid{0000-0002-0337-6877}
\author{Kangcheng Wu}
\orcid{***}
\author{Junjie Chen}
\orcid{***}
\author{Ziwen Tan}
\orcid{***}
\author{Yuming Fang}
\orcid{***}
\email{yanjiebin@jxufe.edu.cn}
\affiliation{%
  \institution{School of Computing and Artificial Intelligence, Jiangxi University of Finance and Economics}
  \city{Nanchang}
  \country{China}
}

\author{Weide Liu}
\email{weide001@e.ntu.edu.sg}
\orcid{***}
\affiliation{%
  \institution{Harvard Medical School, Harvard University}
  \city{Boston}
  \country{USA}
}

\authorsaddresses{
Authors' addresses: J. Yan, K. Wu, J. Chen, Z. Tan, Y. Fang (Corresponding author), Jiangxi University of Finance and Economics, Nanchang, China; emails: yanjiebin@jxufe.edu.cn, kangchengwu-my@foxmail.com, chen.bys@outlook.com, ziwentan@foxmail.com, fa0001ng@e.ntu.edu.sg; W. Liu, Harvard Medical School, Harvard University, USA; email: weide001@e.ntu.edu.sg.
}

\renewcommand{\shortauthors}{J Yan et al.}

\begin{abstract}
Most of existing blind omnidirectional image quality assessment (BOIQA) models rely on viewport generation by modeling user viewing behavior or transforming omnidirectional images (OIs) into varying formats; however, these methods are either computationally expensive or less scalable. To solve these issues, in this paper, we present a flexible and effective paradigm, which is \textit{viewport-unaware} and can be easily adapted to 2D plane image quality assessment (2D-IQA). Specifically, the proposed BOIQA model includes an adaptive prior-equator sampling module for extracting a patch sequence from the equirectangular projection (ERP) image in a resolution-agnostic manner, a progressive deformation-unaware feature fusion module which is able to capture patch-wise quality degradation in a deformation-immune way, and a local-to-global quality aggregation module to adaptively map local perception to global quality. Extensive experiments across four OIQA databases (including uniformly distorted OIs and non-uniformly distorted OIs) demonstrate that the proposed model achieves competitive performance with low complexity against other state-of-the-art models, and we also verify its adaptive capacity to 2D-IQA. The source code is available at \url{https://github.com/KangchengWu/OIQA}.
\end{abstract}





\maketitle

\section{Introduction}
\label{sec:intro}

With the rapid advancement of virtual reality (VR) technology, omnidirectional image (OI), also known as VR image, panoramic image, or 360$^\circ$ image, has emerged as an important visual medium. Unlike 2D planar images, OIs are usually displayed in a sphere, allowing end-users to freely view 360$^{\circ}$ image content in a virtual environment through head-mounted devices (HMDs), and obtain a realistic and immersive visual experience. Generally, OIs may encounter quality degradation in the entire processing chain, including generation, processing, transmission, and storage~\cite{wang2017begin,fang2017no,fang2017objective,yan2020no}, which severely affects the user's quality of experience (QoE)~\cite{fang2020perceptual}. Therefore, accurate evaluation of the quality of OIs is essential to ensure that users enjoy a high QoE and improve the efficiency of image processing algorithms~\cite{fang2021superpixel,ding2021comparison}.

\begin{figure}[ht]
\centering
\includegraphics[width=1\linewidth]{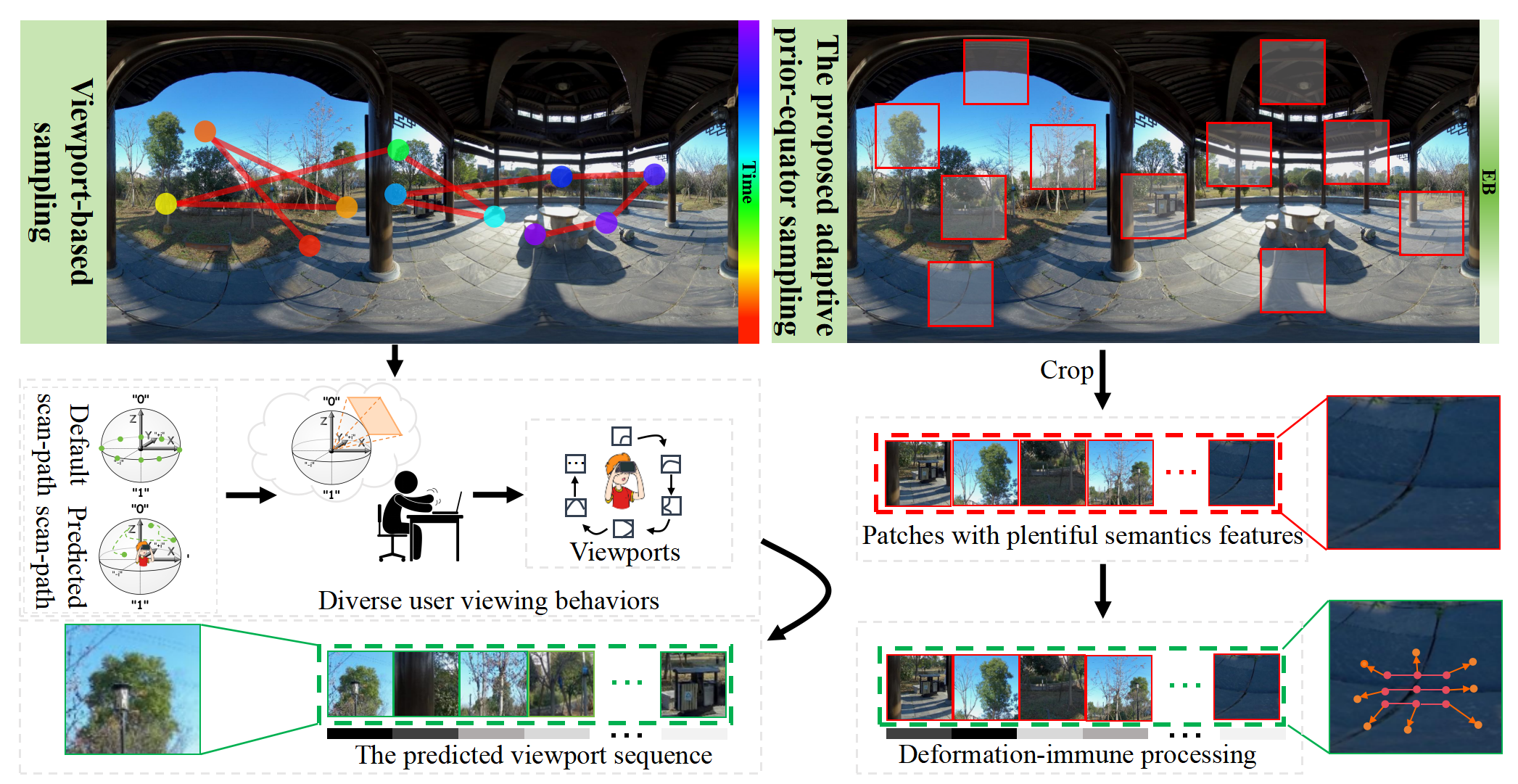}
\caption{An intuitive comparison of the general OI processing procedure (left) and the proposed one (right). (left) Researchers usually extract viewports using a ``computationally expensive'' scan-path prediction method or relying on the ``often unavailable'' ground-truth viewing behavior; (right) We simplify this procedure substituted by extracting patches from the ``visually deformed'' OI directly.}
\label{fig:motivation}
\end{figure}

Omnidirectional image quality assessment (OIQA) can be classified into subjective and objective OIQA, where the former means that subjective participants view OIs and then rate the quality based on their perceptions~\cite{yan2022subjective}. In this labor intensive way, we can thoroughly investigate the effect of some influencing factors on QoE and establish subjective databases to benchmark objective models. The latter involves designing computational models~\cite{yan2022subjective,fang2022perceptual}, which can be classified into three categories: full-reference OIQA (FR-OIQA), reduced-reference (RR-OIQA), and no-reference/blind OIQA (BOIQA). Since reference images in many real scenarios are unaccessible, BOIQA models are therefore more practical than the other two types of models, \ie, FR-OIQA and RR-OIQA.

Generally, before processing OIs, OIs are often converted to other formats, such as the equirectangular projection (ERP)~\cite{zhou360degree}, cubemap projection (CMP)~\cite{jiang2021cubemap}, pyramid projection (PYM)~\cite{li2023mfan}, and pseudo-cylindrical projection~\cite{li2021pseudocylindrical}. The most common format is ERP, which maps an OI from a spherical form to a plane. However, during this mapping process, the image exhibits systematic distortion, leading to a denser pixel distribution in the equatorial region and a sparser distribution in two polar regions. Therefore, the current popular design paradigm is first to extract viewports from each OI by simulating user viewing behavior, \ie, only a limited field of view can be seen by users at a moment, and then to measure overall quality by aggregating viewport-wise quality scores. As shown in Fig.~\ref{fig:motivation}, most existing BOIQA models~\cite{li2019viewport,xu2019quality,xu2020vgcn} resort themselves to a computationally expensive scan-path prediction model~\cite{sui2023scandmm} or the often unavailable ground-truth viewing behavior~\cite{fang2022perceptual} to get the potential viewports, which are then regarded as input. Admittedly, these operations help BOIQA models more align with human behavior and achieve much success in advancing BOIQA. However, predicting which areas attract the most attention from users is absolutely not an easy task, as it involves many influencing factors, such as image content, distortion distribution, viewing starting point, viewing time, \emph{etc}. This type of architectures are more suitable to \textit{personalized} BOIQA~\cite{sui2021perceptual,fang2022perceptual,liu2024perceptual} due to the diverse viewing behaviors, \emph{i.e.}, one may have different viewing trajectories under distinct viewing conditions, and different users may also have different scan-paths under the same viewing condition. Furthermore, most viewport-aware BOIQA models are verified on small-scale OIQA databases with simple contents and uniformly distributed distortion and suffer an evident decline on large-scale OIQA databases with complex contents and non-uniformly distributed distortion~\cite{yan2024subjective,yan2024omnidirectional} (whose quantitative results are shown in the \textbf{Experiments Section}), \ie, their generalization ability is severely affected by their settings.

To address the aforementioned questions, we propose a flexible and effective paradigm from an alternative perspective, \ie, the holistic view, aiming to accurately predict global quality of OIs without reliance on viewport generation, and name the proposed model Viewport-Unaware BOIQA (VU-BOIQA). Specifically, we start with an adaptive prior-equator sampling (APS) module for extracting patches from ERP images, which allows self-adjustment of patch size according to the resolution of OIs and thus ensuring effective semantic content extraction. Then, we use a backbone equipped with a progressive deformation-unaware feature fusion (PDFF) module, which incorporates deformable convolutions to handle the irregular irrugeometric deformation in ERP images and leverages dynamic inter-layer interactions to achieve effective feature integration from different scales and layers. Finally, global quality is obtained by using a local-to-global quality aggregation (LGQA) module.

In summary, our contributions are as follows:
\begin{itemize}
\item We propose a flexible and effective paradigm, which can avoid the obligatory viewport generation operation in most existing studies, \ie, directly extract a patch sequence from an ERP OI.

\item To accommodate the new paradigm, we design an APS module which can self-adjust the sampling region and size of the image patches to retain semantic information, and introduce a PDFF module to help the proposed model to be resistant to the inborn deformation in ERP format and enhance feature interaction.

\item We conduct comprehensive experiments on four OIQA databases, where VU-BOIQA shows competitive performance with low computational cost against state-of-the-art OIQA models. Additionally, we verify its effectiveness on 2D-IQA databases. 
\end{itemize}

\section{Related Work}\label{sec:related_work}
In this section, we first briefly introduce existing viewport-unaware OIQA models. Then, we describe viewport-based OIQA models and deformable convolution.

\subsection{Viewport-Unaware OIQA} 

A simple way is to extend the traditional 2D-IQA methods~\cite{wang2004image} by considering the spherical characteristics of OIs. Yu~\et~\cite{yu2015framework} proposed S-PSNR, which calculates the PSNR by uniformly sampling pixels on the sphere, ensuring an even distribution of errors. Sun~\et~\cite{sun2017weighted} introduced WS-PSNR, which computes PSNR by weighting different areas on the sphere to better mimic the visual preference of the equatorial regions. Zakharchenko~\et~\cite{zakharchenko2016quality} developed CPP-PSNR, which assesses image quality by projecting OIs in Craster's parabolic projection format. Considering the sensitivity of the human visual system (HVS) to structure information, Chen~\et~\cite{chen2018spherical} and Zhou~\et~\cite{zhou2018weighted} proposed S-SSIM and WS-SSIM respectively. These methods are similar to S-PSNR and WS-PSNR as they use uniform sampling on the sphere to calculate errors and weight different regions.

In the current deep learning era, these traditional methods are far from satisfactory. Lim~\et~\cite{lim2018vr} proposed a deep BOIQA method based on adversarial learning, which estimates the weight of each image patch and the corresponding quality score, and then aggregates these patch-wise scores using their weights. Kim~\et~\cite{kim2019deep} performed local partitioning of ERP images into image patches and then encoded each patch's position and visual features to acquire perceptual quality information. Jiang~\et~\cite{jiang2021cubemap} proposed an OIQA method based on human visual behavior. It uses the six faces of an OI in cubemap format to extract quality features and includes attention feature matrices and subsets to better align with human attention patterns. Subsequently, the quality score is obtained using three different schemes. Zheng~\et~\cite{zheng2020segmented} proposed to convert an OI into the segmented spherical projection (SSP) format to extract features, and then the overall quality score is predicted by perceptual features of bipolar and equatorial regions.  Li~\et~\cite{li2023mfan} proposed a multi-projection fusion attention network for OIQA, which uses an attention module to fine-tune the Contrastive Language-Image Pre-Training (CLIP) features adapting to the OIQA task, applies a multi-layer to predict quality score in each projection space (including CMP, PYM and ERP), and uses the polynomial regression to obtain the final score. Zhou~\et~\cite{zhou360degree} proposed a perception-oriented U-Shaped transformer network, which utilizes multiple CMP image patches with different view directions as input, introduces a salience map to generate the weight of these view discretions and uses a U-Shaped transformer encoder and regression module to obtain the quality score. Liu~\et~\cite{liu2023toward} proposed a BOIQA method that combines gradient-based global structural features with gray-level co-occurrence matrix-based local structural features. The viewport-unaware OIQA methods above process OIs in one or multiple projection spaces; however, their performance is still far from promising, and the deformation problem in the ERP format has not been considered.

\subsection{Viewport-Aware OIQA} This type of models extract a viewport sequence as input, which is more consistent with user viewing behavior. Sun~\et~\cite{sun2019mc360iqa} proposed a multi-channel convolutional neural network (CNN), which extracts six viewports (top, down, left, right, front, and back) given a starting point and enhances feature extraction by adjusting the starting point, and then uses the features from all viewports to predict the final quality. Zhou~\et~\cite{zhou2021omnidirectional} proposed a distortion-discriminative multi-stream neural network by adding a distortion discrimination task for joint optimization, where the features are extracted through a shared network and the quality is estimated by a quality regression network. Xu~\et~\cite{xu2020vgcn} proposed a viewport-oriented graph CNN, which establishes a spatial viewport graph through the local branch to simulate dependencies between viewports, and its output features are fused with the features obtained from the global branch to obtain the final quality score. Zhang~\et~\cite{zhang2022no} considered the impact of scene on perceptual quality and dependency between viewports, and proposed a joint quality perception network, which uses CNN to learn local distortion features and uses a recurrent neural network (RNN) to assign weights to different viewports. Fang~\et~\cite{fang2022perceptual} proposed a BOIQA model which injects viewing conditions in multiscale feature fusion. Wu~\et~\cite{wu2024assessor360} proposed an OIQA network, which comprises a joint recursive viewport sampling module to simulate user viewing behavior, a multiscale feature aggregation module to extract multiscale features of the sampled viewports, and a temporal modeling module to establish correlations between the viewports. Liu~\et~\cite{liu2024perceptual} proposed a transformer-based OIQA model, which incorporates viewing conditions and behaviors to ensure alignment with the human viewing process. It contains a multi-scale feature extraction module and a perceptual quality prediction module. Although these viewport-aware OIQA methods can achieve satisfactory performance, they heavily rely on the computationally expensive viewport sequence prediction module and thus are less practical and flexible in real applications, which is, however, no longer an issue for the proposed model.

\subsection{Deformable Convolution} 
Traditional CNNs face limitations in handling geometric deformations of images due to their fixed convolutional kernel structure. Dai~\et~\cite{dai2017deformable} proposed deformable convolutions (DCN), which integrates offsets into the convolutional kernel, enabling the convolutional kernel to dynamically reposition itself to conform to actual deformations in input features. Zhu~\et~\cite{zhu2019deformable} proposed deformable convolutions v2 (DCNv2), which introduces modulation scalars between the offsets and the convolutional weights to adjust the contribution of each offset position in the convolution operation. This further enhances the model's precision and selectivity in responding to key features, allowing it to more accurately capture and process features that significantly impact the task while ignoring non-critical information. Wang~\et~\cite{wang2023internimage} proposed deformable convolutions v3 (DCNv3), which refines the structure of DCNv2 by dividing its weights into depth-wise and point-wise parts. This division operation helps reduce the number of parameters and complexity. Besides, DCNv3 introduces a multi-group mechanism that allows it to extract more nuanced information from different representational subspaces across various positions. Furthermore, DCNv3 improves the training stability across scales by normalizing the modulation scalars based on sampling points, enhancing overall model effectiveness. Chai~\et~\cite{chai2021monocular} proposed a three-channel network featuring left, right, and binocular disparity views to simulate the interaction of monocular and binocular vision, which replaces standard convolutions with deformable convolutions to ensure consistent receptive fields in ERP images.

\begin{figure*}[t]
\begin{center}
\includegraphics[scale=0.18]{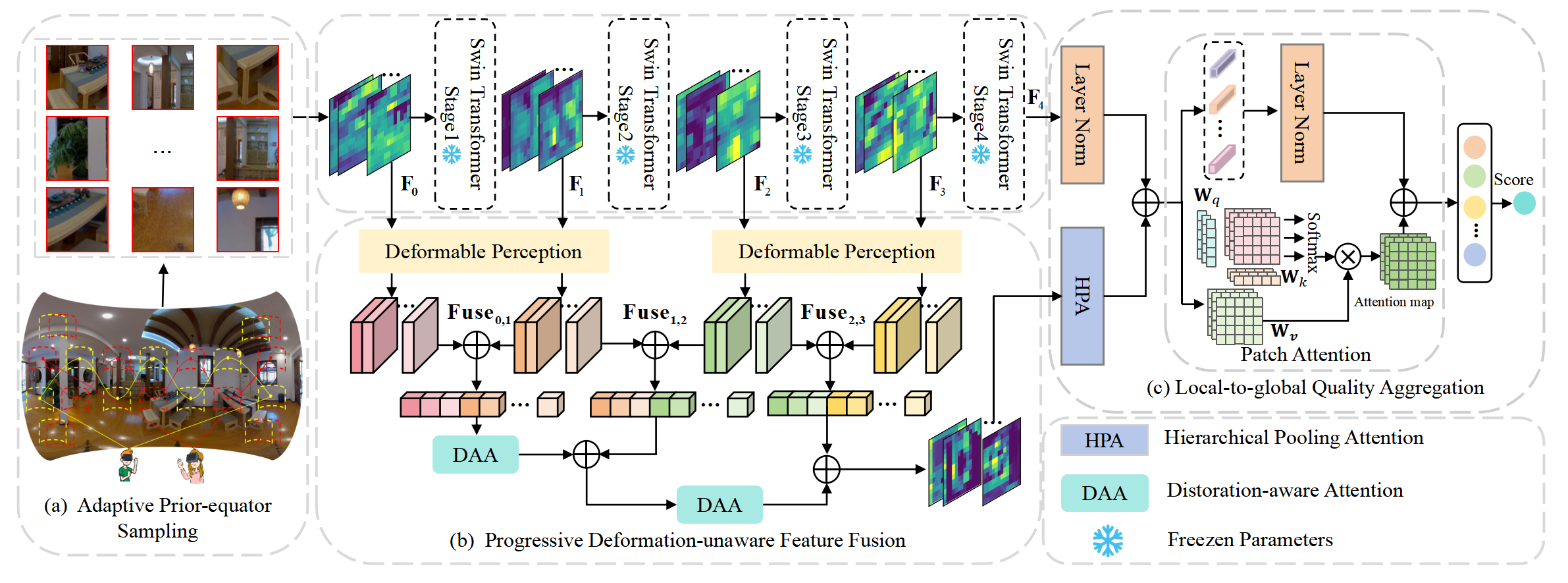}
\end{center}
\caption{The framework of the proposed VU-BOIQA model, which mainly consists of three modules, \ie, an APS module for directly extracting patches from OIs in the ERP format, a PDFF module for jointly addressing the inborn irregular geometric distortion in ERP images and fusing multilevel features, and a LGQA module for mapping patch-wise quality to global quality.}
\label{fig: framework}
\end{figure*}

\section{The Proposed Method}
\subsection{Overview} 
\label{Overview}
To address the aforementioned issue, we design a novel BOIQA method with the pursuit of generalization and flexibility,~\ie, without any projection operations or computationally expensive viewport generation process, and aligns with human visual perception. The overall framework, as depicted in Fig.~\ref{fig: framework}, consists of three key modules, including the APS module (which differentiates it from viewport-based BOIQA models), the PDFF module (designed to capture raw patch-wise degradation), and the LGQA module (which adaptively fuses patch-wise quality scores). We then describe each of these modules one by one in detail.

\begin{figure}[ht]
\centering
\includegraphics[width=0.7\linewidth]{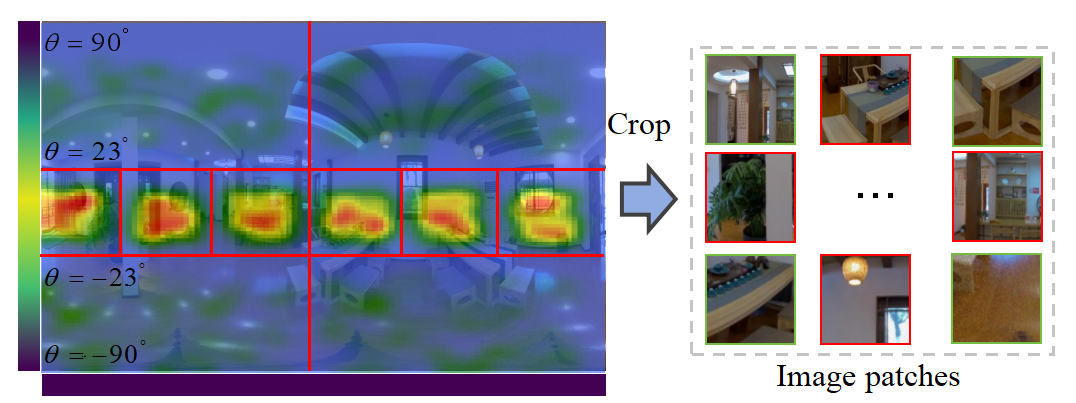}
\caption{The diagram of the proposed APS module, which performs random sampling in both latitude and longitude directions based on the equatorial prior distribution.}
\label{fig:aps}
\end{figure}

\begin{figure}[ht]
\begin{center}
\includegraphics[scale=0.19]{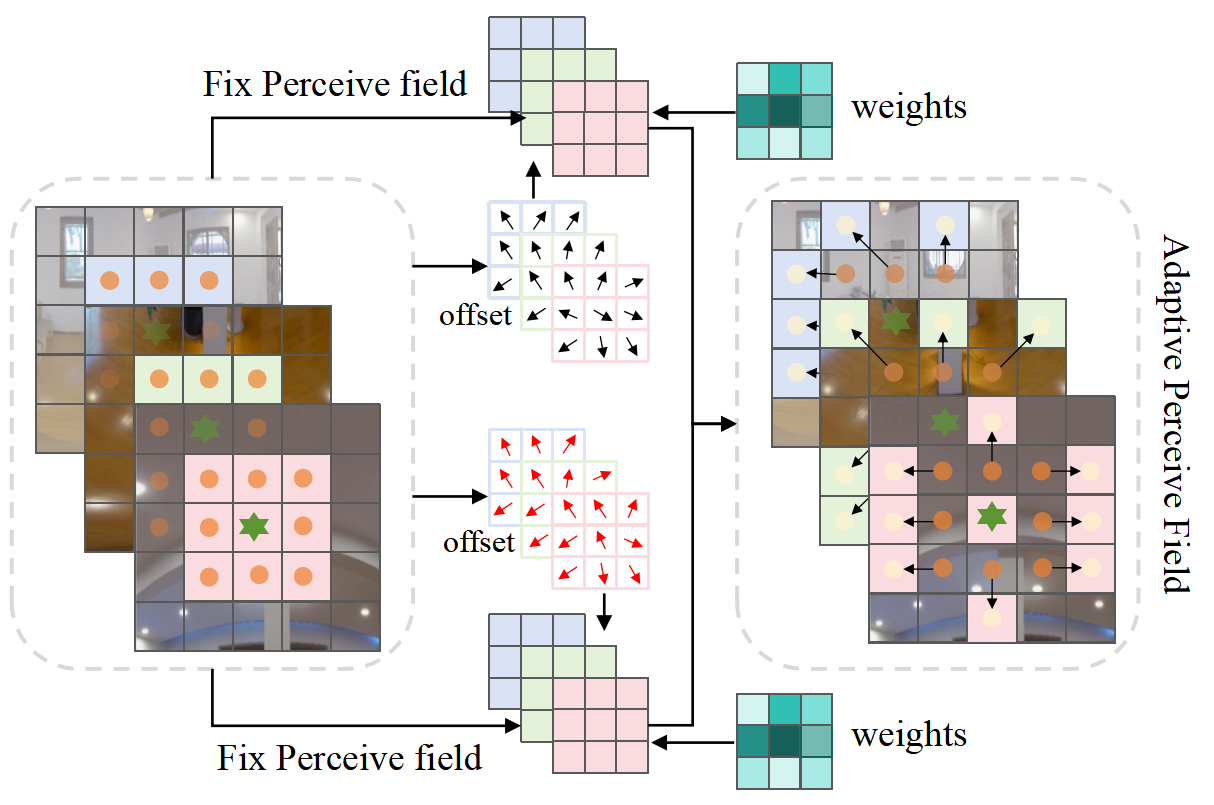}
\end{center}
\caption{The deformation perception operation generates offsets for each sampling position through convolutional layers, dynamically adjusts the convolutional kernel’s sampling locations based on these offsets, and performs weighted aggregation of the sampled features.}
\label{fig: adaptive perceive}
\end{figure}

\subsection{Adaptive Prior-equator Sampling Module} 
The APS module introduces a novel approach by directly extracting patches from an ERP image, as illustrated in Fig.~\ref{fig:aps}. This eliminates the reliance on viewports and significantly reduces both the time-consuming and the complexity associated with viewport extraction. Considering that semantic information in OIs becomes sparse near two polar regions and users usually pay more attention to the regions near the equator, which aligns with an equatorial prior distribution in latitude and uniform distribution in longitude, we sample patches based on the prior probabilities. Given a pixel coordinates $(x,y)$, the coordinate $X$ follows a prior equator (PE) distribution $PE(\mu, \lambda)$ and $Y$ follows a uniform distribution $U(-180, 180)$, where $\mu$ and $\lambda$ are the location and scale parameters, respectively. Usually, we set $\mu = 91.3^\circ$ and $\lambda = 18.58^\circ$. We empirically set the latitude division factor $\theta_T = 23$ and the number of image patches to $K$. The low and high latitude regions can be respectively divided into the following number of equal distance grid blocks:
\begin{equation}
  \begin{aligned}
    &K_{\textit{low}} = K \times \text{Floor}\left[\int_{-\theta_T}^{\theta_T} f_{\text{PE}}(x) \, dx\right], \\
    &K_{\textit{high}} = K - K_{\textit{low}},
  \end{aligned}
\end{equation}
where $f_\text{PE}$ is the function of following $PE$ probability distribution, and $K_\textit{low}$ and $K_\textit{high}$ are the numbers of grid blocks in low latitude and high latitude, respectively. Due to symmetry along the latitude, the northern and southern regions are divided into $K_{high}$ grid blocks. We then randomly sample each grid block in latitude and longitude directions. The size of the sampling blocks can be determined as follows:
\begin{equation}
    \begin{aligned}
      &P_w = E_w \times \kappa_w,\\
      &P_h = E_h \times \kappa_h,  
    \end{aligned}
\end{equation}
where $P_w$ and $P_h$ denote the width and height of image patches, $E_w$ and $E_h$ is the width and height of ERP image, $\kappa_w$ and $\kappa_h$ are the sampling factor of width and height, which is used to adaptively adjust the patch size according to the resolution of OIs for sampling more effective semantic information. Subsequently, we perform random sampling within each grid block in both latitude and longitude directions, obtaining a set of coordinates of image patches, denoted by $\{(x_1,y_1),(x_2,y_2),...,(x_K,y_K)\}$. We crop the ERP image according to the coordinates and patch size to obtain a collection of image patches $\textbf{I} \in \{\textbf{I}_1, \textbf{I}_2,..., \textbf{I}_K\}$. The above operation aims to extract more image patches that contain adequate semantic information and align with user viewing behavior. 

\begin{figure}[t]
\begin{center}
\includegraphics[scale=0.20]{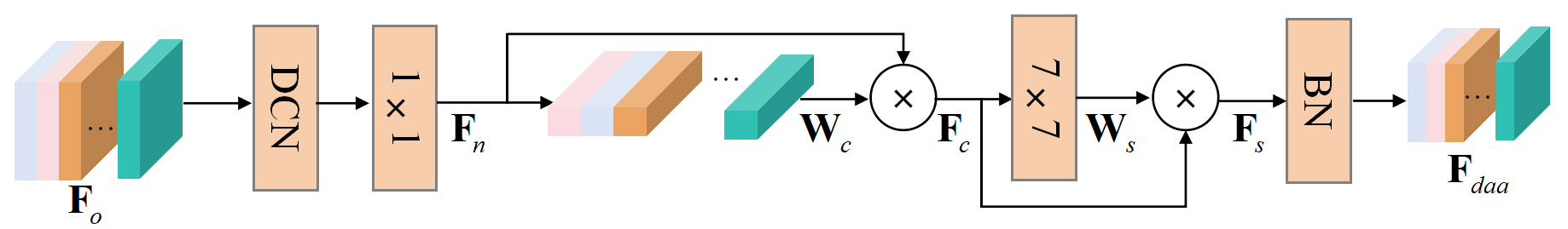}
\end{center}
\caption{The architecture of the DAA module.}
\label{fig: daa}
\end{figure}

\subsection{Progressive Deformation-unaware Feature Fusion} 

Due to its hierarchical structure and efficient representation ability, the Swin Transformer V2~\cite{liu2022swin} demonstrates substantial advantages in various fields such as medicine~\cite{liu2024transdiff}, segmentation~\cite{ye2024segment} and human action recognition~\cite{bian2024}. Here, we select the Swin transformer V2 as the backbone (denoted by $B$). To preserve the original feature extraction capability of $B$ while allowing the network to focus more on recognizing irregular distortion features during the learning process, we freeze the parameters of $B$. Subsequently, the $B$ is used to extract cross layer features, which can be formulated as follows:
\begin{equation}
 \mathbf{F}_i = \{B(\textbf{I};\theta_B)\}_{i=0}^4,
\end{equation}
where $\mathbf{F}_i$ represents the feature map from the $i$-th stage and patch partition of $B$, and $\theta_B$ denotes its learnable parameters. Since the original ERP image feature maps contain many irregular distortions, it is difficult for the general network to capture these details. To address this issue, we adopt the DCN~\cite{wang2023internimage} to adjust these feature maps as follows: 
\begin{equation}
  \mathbf{F}_d^i = \{\text{DCN}(\mathbf{F}_i)\}_{i=0}^3,
\end{equation}
where $\mathbf{F}_d^i$ represents the $i$-th feature map adjusted by $\text{DCN}$. Considering that the kernel of the DCN can adaptively adjust convolution kernel positions (can refer to Fig.~\ref{fig: adaptive perceive}), it is able to effectively handle scale variations in images, bringing better ability in fusing features from different scales. To balance the relationship between local details and the global context of features, these features are then added with adjacent features in an element-wise way as follows: 
\begin{equation}
  \begin{aligned}
    &\mathbf{Fuse}_{i,i+1} = \text{Conv}_{1 \times 1}(\mathbf{F}_d^i) {\oplus} \mathbf{F}_d^{i+1}, i=0,1,2, \\
    &\mathbf{Fuse}_{i,i+1}^\textit{d} = \text{DCN}(\mathbf{Fuse}_{i,i+1}),i=1,2,
  \end{aligned}
\end{equation}
where $\mathbf{Fuse_{i,i+1}}$ denotes the fused feature of $i$-th and $(i+1)$-th, $\text{Conv}_{1 \times 1}$ represents the 2D convolution with $1 \times 1$ kernel and $stride=2$, ${\oplus}$ is the element-wise addition operation. $\mathbf{Fuse_{i,i+1}^\textit{d}}$ denotes the further processed feature map by Wang \emph{et al.}~\cite{wang2023internimage}. To extract high-level semantic features while retaining low-level features and incorporating more complex expressions, we enhance the features fusion by introducing a distortion-aware attention (DAA) module, as shown in Fig.~\ref{fig: daa}. This module first introduces the DCNv3, which is able to dynamically adjust the sampling positions of the convolutional kernels, thereby enhancing the ability to capture irregular distortion. Before the DCN, the DAA module utilizes a 1x1 convolution to adjust inputting channels, which can be formulated as follows:
\begin{equation}
\mathbf{F}_{n} = \text{Conv}_{1 \times 1}(\text{DCN}(\mathbf{F}_{o})),
\end{equation}
where $\mathbf{F}_{n}$ denotes the output feature maps, the $\mathbf{F}_{o}$ is the input feature maps send in DAA module. Subsequently, the module introduces channel and spatial attention mechanism~\cite{woo2018cbam} for adaptively assigning weights to different channels and helping the network focus on the most representative feature channels as follows:
\begin{equation}
    \mathbf{F}_c =\mathbf{F}_n \times (\text{FC}_2(\text{FC}_1(\text{Avgpool}(\mathbf{F}_n)))),
\end{equation}
where $\text{FC}_1$ and $\text{FC}_2$ represent fully connected (FC) layer, $\mathbf{F}_c$ is the feature map refined by the channel attention. Additionally, it applies weighting to the distortion regions in the spatial dimension, significantly enhancing the network's ability to perceive distortion areas in the image as follows:
    \begin{equation}
  \mathbf{F}_{daa} =\text{Maxpool}(\mathbf{F}_c \times (\text{Conv}_{7 \times 7}(\mathbf{F}_c))),
\end{equation}
where $\mathbf{F}_{daa}$ denotes the output feature map of DAA module, $\text{Conv}_{7 \times 7}$ represents the 2D convolution with $\textit{ kernel size}$ = 7, $\textit{output channels}$ = 1 and $\textit{stride}$ = 3, $\text{Maxpool}$ represents the Max pooling Layer with $stride$=2 and $\textit{ kernel size}$ = 2. Lastly, the feature map of the PDFF module was obtained by the formulation: 
\begin{equation}
  \mathbf{F}_{m} = \text{DAA}(\text{DAA}(\mathbf{Fuse}_{0,1}){\oplus}\mathbf{Fuse}_{1,2}^d{\oplus}\mathbf{Fuse}_{2,3}^d),
\end{equation}
where $\mathbf{F}_{m}$ represents the output features of the PDFF module.

\begin{figure}[t]
\begin{center}
\includegraphics[scale=0.19]{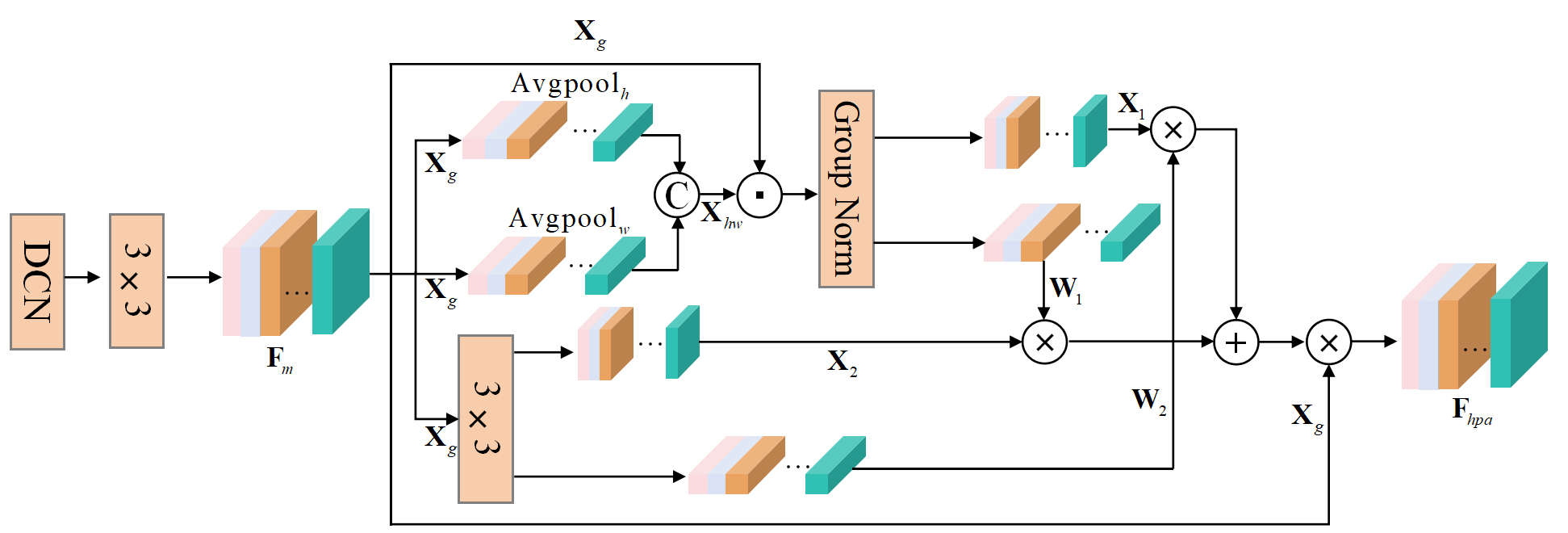}
\end{center}
\caption{The architecture of the HPA module.}
\label{fig: hpa}
\end{figure}

\subsection{Local-to-global Quality Aggregation}

We utilize the HPA module, as shown in Fig.~\ref{fig: hpa}, which integrates DCN, hierarchical pooling, and convolution operations, complemented by an attention mechanism. This combination effectively extracts and enhances spatial features. This module dynamically captures spatial correlations across dimensions, improving the model's ability to detect and extract features from distorted regions. First, we use DCN and $3 \times 3$ convolution to further enhance the local spatial features from the output feature maps of the PDFF module as follows:
\begin{equation}
    \mathbf{F}_{a} = \text{Conv}_{3 \times 3}(\text{DCN}(\mathbf{F}_{m})),
\end{equation}
where $\mathbf{F}_{a}$ denotes the feature map after enhancing, $\text{Conv}_{3 \times 3}$ represents 2D convolution with $kernel size = 3\times3$. The input feature map is divided into multiple groups by channels, enabling group-wise processing to reduce computational complexity and more effectively capture local features as follows:
\begin{equation}
    \mathbf{X}_{g} = \text{Reshape}(\mathbf{F}_{a}),
\end{equation}
where $\mathbf{X}_{g}$ denote grouping feature, $\mathbf{X}_{g} \in \mathbb{R}^{b \times \textit{group} \times \textit{batchsize} \times h \times w}$. To capture global features along the height and width dimensions, we apply adaptive average pooling to the grouped features separately in the height and width directions, and then concatenate them as follows:
\begin{equation}
   \mathbf{X}_{w}, \mathbf{X}_{h} = \text{Split}(\text{Concat}(\text{Avgpool}_h(\mathbf{X}_{g}), \text{Avgpool}_w(\mathbf{X}_{g})) \odot \mathbf{X}_{g}),
\end{equation}
where $X_w$ and $X_h$ denote the feature maps along the width and height dimensions, respectively, $\text{Split}$ represents the splitting operation alone width and height dimensions, $\text{Concat}$ represents the concatenation operation, which is performed along dimension 2, the $\text{Avgpool}_h$ and $\text{Avgpool}_w$ is average pooling along height and width dimentions, $\odot$ is dot multiply. Subsequently, the grouped features are normalized along the channel dimension to generate global features, followed by a global average pooling and the softmax function to generate normalized attention weights. The whole process can be described as:
\begin{equation}
  \begin{aligned}
    &\mathbf{X}_{1} = \text{GN}(\mathbf{X}_{g} \times \text{Sigmoid}(\mathbf{X}_{h}) \times \text{Sigmoid}(\mathbf{X}_w^T)),  \\
    &\mathbf{W}_1 = \text{Softmax}(\text{Avgpool}(\mathbf{X}_1)),
  \end{aligned}
\end{equation}
where $\mathbf{X}_{1}$ denotes the output feature maps weighted by the height and width attention mechanism, $\text{GN}$ represents Group Normalization, $\mathbf{X}_w^T$ is $\mathbf{X}_w$ after transpose, $\mathbf{W}_1$ represents weight matrix. By applying convolution operations to the originally grouped feature map, local spatial features are further extracted, followed by average pooling and softmax to obtain the weight matrix. The process can be formulated as:
\begin{equation}
  \begin{aligned}
    &\mathbf{X}_{2} =\text{Conv}_{3 \times 3}(\mathbf{X}_{g}),\\
    &\mathbf{W}_2 = \text{Softmax}(\text{Avgpool}(\mathbf{X}_2)),
  \end{aligned}
\end{equation}
where $\mathbf{X}_{2}$ denotes the feature mapes after enhanced by 2D convolution, $\mathbf{W}_2$ denotes weight matrix, $\text{Softmax}$ denotes the softmax function, $\text{Avgpool}$ is average pooling. Finally, the correlation between features is computed through matrix multiplication to generate the final attention weight matrix, which is then multiplied with the initial feature map to obtain the feature map processed by the HPA module as follows:
\begin{equation}
    \mathbf{F}_{hpa} = (\mathbf{X}_1 \times \mathbf{W}_2 + \mathbf{X}_2 \times \mathbf{W}_1) \times \mathbf{X}_g,
\end{equation}
where $\mathbf{F}_{hpa}$ denotes the final output feature map of HPA module. By fusing pooled low-level and high-level features, the model's output retains local detail from lower layers while effectively incorporating global semantics from higher layers. This feature fusion not only enhances the model's feature omnidirectional quality perceive capacity but also significantly reduces computational overhead, the formulation as: 
\begin{equation}
    \mathbf{V}_i = \text{Avgpool}(\mathbf{F}_4) + \text{Avgpool}(\mathbf{F}_{\textit{hpa}}),
\end{equation}
where $\mathbf{V}_i$ denotes feature vector of $i$-th image patch after fusing. $\mathbf{F}_4$ is the output feature of $B$. After concatenating the feature vectors of the ten images, we integrate the patch attention (PA) module with the multi-head self-attention mechanism and feed-forward neural networks to construct a distortion-aware contextual feature representation. This representation is designed to enhance the interaction between local and global information, enabling the model to selectively focus on relevant image patches. By doing so, it effectively captures spatial dependencies between patches while filtering out redundant or irrelevant information. First, we transform the input feature vectors into a higher embedding dimension, then add positional information to each feature vector to help the model learn the spatial relationships between different image patches. The process can be described as:
\begin{equation}
\mathbf{X}_{p} = (\mathbf{W}_{e} \cdot \mathbf{V}_{c} + \mathbf{b}_{e})+\mathbf{P},
\end{equation}
where $\mathbf{X}_{p}$ represents the feature vector obtained after adding the embedding layer and positional encoding, $\mathbf{W}_{e}$ represents the weight matrix of the embedding layer, $\mathbf{V}_{c}$ denotes the feature vector obtained by concatenating the feature vectors of different image patches, $\mathbf{b}_{e}$ represents the bias of the embedding layer, $\mathbf{P}$ is positional encoding matrix, which can be described as:
\begin{equation}
    \mathbf{P} = \text{Parameter}(\text{Randn}(1, l_p, l_d)),
\end{equation}
where \text{Randn ()} denotes random function, which generates a tensor with random values drawn from a standard normal distribution, \text{Parameter ()} is parameterized function, $l_p$ denotes the number of patches, $l_d$ represents the embedding dimension. Subsequently, multi-head self-attention mechanisms assign different attention weights to each image patch, effectively evaluating the importance of each patch. The whole process can be formulated by:
\begin{equation}
  \begin{aligned}
    &\mathbf{V}_m = \text{Concat}(\mathbf{Head}_1, \dots, \mathbf{Head}_i) \mathbf{W}_o,\\
    &\mathbf{Head}_i = \text{Softmax}\left( \frac{\mathbf{X}_p \cdot \mathbf{W}_i^Q \cdot (\mathbf{X}_p \cdot \mathbf{W}_i^K)^T}{\sqrt{d_k}} \right) \mathbf{X}_p \cdot \mathbf{W}_i^V,
  \end{aligned}
\end{equation}
where $\mathbf{V}_m$ represents the feature vector after multi-head self-attention weight allocation, $\mathbf{Head}_i$ denotes the $i$-th attention head, $W_i^Q$, $W_i^K$, and $W_i^V$ are the weight matrices associated with the $i$-th attention head. $d_k$ denotes the dimension of the query (Q) or key (K), and the range of $i$ is from 1 to 8. We further enhance the model's ability to perceive features in distorted regions through the feed-forward neural network, which is obtained by:
\begin{equation}
\mathbf{V}_o = \text{LN}_2 \left( \text{FC} \left( \text{LN}_1(\mathbf{V}_m + \mathbf{X}_p) \right) + \text{LN}_1(\mathbf{V}_m + \mathbf{X}_p) \right),
\end{equation}
where $\mathbf{V}_o$ represents the final output feature vector, $\text{LN}_1$ and $\text{LN}_2$ denote layer normalization operations. The final perceptual quality score of the image is obtained by taking the average of ten feature vectors after they have been processed through a FC layer:

\begin{equation}
 \hat{s}_i = \frac{1}{K}(\mathbf{FC}(\mathbf{V}_o, \theta_f)),
\end{equation}
where ${s}_i$ denotes the final perceptual quality score of $i-$th OI, $\theta_f$ denotes the parameters of the FC layer. In this study, we use norm-in-norm loss \cite{li2020norm} function to train and optimize the model, which is defined as:

\begin{equation}
  \textit{L}(s_n, \hat{s}_n) = \frac{1}{\omega N} \sum_{n=1}^N \left\| \frac{s_n - \text{M}(s)}{\sigma} - \frac{{\hat{s}}_n - \text{M}(\hat{s})}{\hat{\sigma}} \right\|^\gamma,
\end{equation}
where $\textit{L}$ denotes the norm-in-norm loss function, $\omega$ is the normalization factor, $\gamma$ is a hyperparameter, $\hat{s}$ is the set of all predicted quality scores, $\hat{s} \in \{\hat{s}_1,\hat{s}_2,....,\hat{s}_N\}$, $S$ is the set of all subjective quality scores, $s \in \{s_1,s_2,....,s_N\}$, $\text{M}$ is mean function, $\sigma$ and $\hat{\sigma}$ represent the variances of the predicted and subjective quality scores, respectively.

\section{Experiments}
\subsection{Test Databases}
\begin{table*}[t]
\setlength{\tabcolsep}{2.8mm}{
\caption{Performance comparison of the proposed VU-BOIQA and state-of-the-art OIQA and 2D-IQA methods. The best two results are highlighted in bold.}
\label{OIQA compare}
\begin{tabular}{@{}ccccccccccc@{}}
\toprule
 &    &    & \multicolumn{2}{c}{CVIQ} & \multicolumn{2}{c}{OIQA} & \multicolumn{2}{c}{JUFE-10K} & \multicolumn{2}{c}{OIQ-10K}  \\ \cmidrule(l){4-11} 
\multirow{-2}{*}{Method} & \multirow{-2}{*}{\begin{tabular}[c]{@{}c@{}}Params\\ (M)\end{tabular}} & \multirow{-2}{*}{\begin{tabular}[c]{@{}c@{}}FLOPs\\ (G)\end{tabular}} & SRCC    & PLCC   & SRCC    & PLCC   & SRCC  & PLCC   & SRCC      & PLCC      \\\midrule
S-PSNR~\cite{yu2015framework}     & N/A    & {N/A}   & 0.708   & 0.708   & 0.599    & 0.539    & 0.355   & 0.285    & 0.302   & 0.251            \\
WS-PSNR~\cite{sun2017weighted}    & N/A    & {N/A}   & 0.672   & 0.610   & 0.581    & 0.526    & 0.353   & 0.284    & 0.295   & 0.248            \\
CPP-PSNR~\cite{zakharchenko2016quality}   & N/A    & {N/A}   & 0.687   & 0.626   & 0.568    & 0.514    & 0.355   & 0.285    & 0.295   & 0.248            \\
WS-SSIM~\cite{zhou2018weighted}    & N/A    & {N/A}   & 0.929   & 0.911   & 0.504    & 0.503    & 0.388   & 0.249    & 0.223   & 0.062            \\\midrule
HyperIQA~\cite{su2020blindly}   & 27.4   & N/A     & 0.603   & 0.597   & 0.606    & 0.577    & 0.201   & 0.198    & 0.470   & 0.443        \\
MANIQA~\cite{yang2022maniqa}     & 135.7  & 108.6   & 0.287   & 0.279   & 0.367    & 0.308    & 0.101   & 0.090    & 0.112   & 0.097           \\
VCRNet~\cite{pan2022vcrnet}     & 16.7   & 10.3    & 0.390   & 0.392   & 0.425    & 0.382    & 0.162   & 0.150    & 0.289   & 0.277            \\
LIQE~\cite{zhang2023blind}       & 151    & 44.2    & 0.484   & 0.535   & 0.767    & 0.873    & 0.047   & 0.026    & 0.551   & 0.521         \\\midrule
Assessor360~\cite{wu2024assessor360} & 88.2    & 230.5   &  \textbf{0.976}     &  \textbf{0.964}   & \textbf{0.974}    & \textbf{0.980} & \textbf{0.694}    & \textbf{0.690}    & \textbf{0.790}  & \textbf{0.773}       \\
MC360IQA~\cite{sun2019mc360iqa}   & 22.4 & 30.3   & 0.950  & 0.913   & 0.924    & 0.918     & 0.620  & 0.611    & 0.721  & 0.710       \\
VGCN~\cite{xu2020vgcn}       & 26.5 & 191.5  & \textbf{0.965}        & \textbf{0.963}   & 0.958   & 0.951         & 0.464    & 0.367  & 0.705 & 0.698                        \\
Fang22~\cite{fang2022perceptual}      & 25.2 & 174.3  & -   & -   & -  & -     & 0.633     & 0.616           & \textbf{0.769}   & \textbf{0.758} \\\midrule
VU-BOIQA        & 30.2 & 40.8   &0.963   & 0.958    & \textbf{0.976} & \textbf{0.973}   & \textbf{0.782} & \textbf{0.781} &  0.766 & 0.749 \\ \bottomrule
\end{tabular}}
\end{table*}
The proposed model is evaluated on four OIQA databases, including CVIQ~\cite{sun2018large}, OIQA~\cite{duan2018perceptual}, JUFE-10K~\cite{yan2024subjective}, and OIQ-10K~\cite{yan2024omnidirectional}. Specifically, the CVIQ database includes 16 reference OIs and 528 compressed OIs. These images have been compressed using three common encoding techniques (JPEG, H.264/AVC, and H.265/HEVC), each of which includes 11 levels of compression severity. The OIQA database consists of 16 reference OIs and 320 uniformly distorted OIs, which are compressed by four common types of distortions (JPEG compression, JPEG2000 compression, Gaussian blur and Gaussian noise). The JUFE-10K database contains 10,320 OIs with non-uniform distortion, and the OIQ-10K database includes 10,000 OIs with both uniform and non-uniform distortion.

\subsection{Evaluation Metrics}
Three performance criteria, including Pearson’s linear correlation coefficient (PLCC), Spearman’s rank order correlation coefficient (SRCC), and root mean square error (RMSE), are used to measure the prediction monotonicity and accuracy. Higher PLCC and SRCC and lower RMSE denote better performance. 
As suggested in~\cite{vqeg}, the predicted scores are first mapped to subjective ratings before calculating PLCC and RMSE. We adopt the following five-parameter logistic function:

\begin{equation}
  f(x) = \rho_1\left(\frac{1}{2} - \frac{1}{1 + e^{\rho_2(x-p_3)}}\right) + \rho_4{x} + \rho_5,
\end{equation}
where $f(x)$ denotes the data after fitting, the $x$ represents the original data, $\rho_1,\rho_2,\rho_3,\rho_4,\rho_5$ are the fitted parameters.

\subsection{Implementation Details}\label{implement details}
\label{Experimental_Setups}
In the experiments, we divide the OIs in each database into 80$\%$ for training and 20$\%$ for testing. For each image, 10 original patches are extracted using APS, with sampling factors set to $\kappa_h=$ 0.2 and $\kappa_w= $ 0.1. These patches are then resized to 224×224 pixels before being inputted into the proposed model. The backbone network of the proposed model is based on Swin Transformer V2, and the weights for this backbone are initialized from pre-trained models on ImageNet 1K~\cite{deng2009imagenet} for transfer learning. This initialization helps the model leverage feature representations learned from large-scale datasets, providing a good starting point for our task. The model is trained on a high-performance server equipped with an Intel(R) Xeon(R) Gold 6326 CPU @ 2.90 GHz, 24 GB NVIDIA GeForce RTX A5000 GPU, and 260 GB RAM. The Adam optimizer is used with an initial learning rate of 1e-4 and the batch size is set to 8. The training process ends after 25 epochs.

\subsection{Performance Comparison}

We compare the proposed model with several state-of-the-art 2D-IQA and OIQA models. The 2D-IQA methods include HyperIQA~\cite{su2020blindly}, MANIQA~\cite{yang2022maniqa}, VCRNet~\cite{pan2022vcrnet}, and LIQE~\cite{zhang2023blind}. We loaded the publicly available weights of these methods and strictly followed the experimental settings outlined in the original papers (e.g., sampling strategy, number of sampled patches). We directly input the image patches from the ERP images into the models for testing. FR-OIQA methods include S-PSNR~\cite{yu2015framework}, WS-PSNR~\cite{sun2017weighted}, CPP-PSNR~\cite{zakharchenko2016quality}, and WS-SSIM~\cite{zhou2018weighted}, and BOIQA methods include Assessor360~\cite{wu2024assessor360}, MC360IQA~\cite{sun2019mc360iqa}, VGCN~\cite{xu2020vgcn}, and Fang22~\cite{fang2022perceptual}, we retrain them on the JUFE-10K and OIQ-10K databases using the same training and test splitting scheme to ensure a fair comparison. The experimental results are shown in Tab.~\ref{OIQA compare}, where the top two results are highlighted. As shown in Tab.~\ref{OIQA compare}, all 2D-IQA models show poor performance on the OIQA databases, which is attributed to the fact that the 2D-IQA models do not consider the spherical characteristic of OIs. Among these 2D-IQA models, HyperIQA performs slightly better than the others. This advantage may be due to HyperIQA's ability to integrate both local distortion features and global semantic features, enabling it to capture non-uniform distortions more effectively. In terms of FR-OIQA methods \emph{i.e.}, S-PSNR, WS-PSNR, CPP-PSNR, and WS-SSIM, we can clearly observe that traditional FR-OIQA models outperform 2D-IQA models. This is reasonable, as FR-OIQA models take into account the geometric characteristics of OIs and are designed based on the PSNR and SSIM metrics. However, it is difficult for them to accurately assess distorted OIs because their manually designed feature extraction cannot adapt well to complex distortion environments.

Compared to these FR-OIQA and 2D-IQA models, the deep learning based BOIQA models show better performance on the OIQA databases due to their specific designs on model architecture. Note that these BOIQA models show worse performance on the JUFE-10K and OIQ-10K databases than on the CVIQ and OIQA databases, which is attributed to that the JUFE-10K and OIQ-10K databases contain uniform and non-uniform distortion, bringing new challenge to OIQA models. As for these advanced deep learning based BOIQA models \ie, Assessor360, MC360IQA, VGCN, Fang22, we can clearly observe that Assessor360 outperforms other models. This may be attributed to its unique Recursive Probability Sampling (RPS) scheme, which enables it to capture more semantic information. Additionally, another possible reason is that its backbone network (\ie, the Swin Transformer) is better at perceiving distortion information in OIs compared to other networks.

The proposed model ranks in the top two on the CVIQ and OIQA databases with a gap of less than 1.12$\%$, and outperforms the best performing Assessor360 by 10.88$\%$ on the JUFE-10K database. On the OIQ-10K dataset, our model does not achieve top-two performance, which may be due to the existence of many high-resolution OIs. As indicated by subsequent experimental results (see Sections \ref{np} and \ref{ssf}), the performance of our model depends on the sampling factor and the number of sampling patches. Therefore, with current settings, it is suboptimal to capture more semantic information in these high-resolution OIs, leading to a decline in performance. However, the performance gap compared to the best model, Assessor360, is only 3.0$\%$. In terms of FLOPs and parameters, our method reduces by 65.7$\%$ and 82$\%$, respectively, compared to Assessor360. Additionally, our proposed method achieves this simply by extracting image patches, without requiring the complex viewport extraction process used in RPS.

\subsection{Generalizability Validation}
In order to validate the generalization ability of the proposed VU-BOIQA, cross-database validation is carried out on the JUFE-10K and OIQ-10K databases. Specifically, we randomly select 80$\%$ images in one database as the training set and test it on the other database. The experimental results are listed in Tab.~\ref{cross validation}. From Tab.~\ref{cross validation}, we can observe that the proposed model achieves competitive performance in cross-database validation, demonstrating its strong generalizability to different databases.
\begin{table}[]
\caption{Cross database validation. The best results are highlighted in bold.}
\setlength{\tabcolsep}{5mm}{
\begin{tabular}{@{}ccccccc@{}}
\toprule
\multirow{3}{*}{Models} & \multicolumn{3}{c}{Train JUFE-10K} & \multicolumn{3}{c}{Train OIQ-10K} \\  
                        & \multicolumn{3}{c}{Test OIQ-10K}   & \multicolumn{3}{c}{Test JUFE-10K} \\\cmidrule(r){2-7}
                        & PLCC       & SRCC      & RMSE      & PLCC      & SRCC      & RMSE      \\ \cmidrule(r){1-7}
MC360IQA~\cite{sun2019mc360iqa}                & 0.290      & 0.278     & 1.063     & 0.319     & 0.253     & 0.576     \\ 
VGCN~\cite{xu2020vgcn}                    & 0.426      & 0.418    & 0.415     & 0.550     & 0.517     & 0.508     \\
Fang22~\cite{fang2022perceptual}                  & 0.274      & 0.162     & 0.441     & 0.429     & 0.366     & 0.549     \\
Assessor360~\cite{wu2024assessor360}             & 0.357      & 0.367     & 0.428     & 0.624     & \textbf{0.614}     & 0.475     \\
VU-BOIQA               & \textbf{0.458}        &   \textbf{0.472}         &   \textbf{0.208}        & \textbf{0.625}         &  0.610       &  \textbf{0.125}        \\ \bottomrule
\end{tabular}}

\label{cross validation}
\end{table}

\subsection{Ablation Studies} 

In this section, we conduct ablation studies to verify the effectiveness of each component in VU-BOIQA.

\begin{table}[]
\caption{The experimental results on the JUFE-10K and OIQ-10K database with different number of sampled patches.}
\label{patches}
\setlength{\tabcolsep}{7mm}{
\begin{tabular}{@{}ccccc@{}}
\toprule
\multirow{2}{*}{Patch number} & \multicolumn{2}{c}{JUFE-10K} & \multicolumn{2}{c}{OIQ-10K} \\ \cmidrule(l){2-5} 
                              & PLCC          & SRCC         & PLCC         & SRCC         \\ \cmidrule(r){1-5}
5                             &    0.731         &   0.726           &     0.747         & 0.727             \\
10                            &  \textbf{0.782}             &    0.781          &  0.766            & 0.749             \\
15                            &   \textbf{0.782}           &      \textbf{0.782}        &     \textbf{0.783}        & \textbf{0.762}             \\ 
20                            &     0.503         &     0.484         &     0.751        &     0.727         \\
\bottomrule
\end{tabular}}
\end{table}

\subsubsection{The number of sampled patches.}\label{np}
We test the sensitivity of the proposed VU-BOIQA to the number of sampled patches, and the experimental results are shown in Tab.~\ref{patches}. When the number of image patches reaches 15, the model performance improves significantly. However, when the patch number increases to 20, the model performance on the JUFE-10K and OIQ-10K databases degrades, which may result from additional noise introduced by more less informative patches and redundancy information, and therefore severely hinders the model performance.

\begin{table}[]
\caption{The experimental results on the JUFE-10K and OIQ-10k database with different sampling scales.}
\label{factor}
\setlength{\tabcolsep}{4.5mm}{
\begin{tabular}{@{}clcccc@{}}
\toprule
\multirow{2}{*}{$\kappa_h$} & \multirow{2}{*}{$\kappa_w$} & \multicolumn{2}{c}{JUFE-10K} & \multicolumn{2}{c}{OIQ-10K} \\ \cmidrule(l){3-6} 
                      &                       & PLCC          & SRCC         & PLCC         & SRCC         \\ \cmidrule(r){1-6}
0.100                 & 0.050                 &   0.762       &     0.762    &   0.749           & 0.722             \\
0.150                 & 0.075                 &     0.767    &    0.765     &  0.764            &  0.740            \\
0.200                 & 0.100                 & \textbf{0.782}         &  \textbf{0.781}       &    \textbf{0.766}           &   \textbf{0.749}            \\
0.250                 & 0.125                 &  0.769        &     0.767    & 0.763             &   \textbf{0.749}           \\ \bottomrule
\end{tabular}}
\end{table}

\begin{table}[h]
\centering
\setlength{\tabcolsep}{4.5mm}
\caption{The performance of VU-BOIQA with different components.}
\label{each component}
\begin{tabular}{@{}cccccccc@{}}
\toprule
\multirow{2}{*}{Backbone} & \multirow{2}{*}{PDFF} & \multirow{2}{*}{HPA} & \multirow{2}{*}{PA} & \multicolumn{2}{c}{JUFE-10K} & \multicolumn{2}{c}{OIQ-10K} \\ \cmidrule(l){5-8} 
                          &                       &                      &                     & PLCC          & SRCC         & PLCC         & SRCC         \\ \midrule
\checkmark                &                       &                      &                     & 0.687         & 0.673        & 0.695        & 0.662        \\
\checkmark                & \checkmark            &                      &                     & 0.760         & 0.757        & 0.751        & 0.732        \\
\checkmark                & \checkmark            & \checkmark           &                     & 0.763         & 0.759        & \textbf{0.769}        & \textbf{0.749}        \\
\checkmark                & \checkmark            & \checkmark           & \checkmark          & \textbf{0.782}& \textbf{0.781}& 0.766        &\textbf{ 0.749}        \\ \bottomrule
\end{tabular}
\end{table}

\subsubsection{Sampling scale factor.}
\label{ssf}
We empirically set the ratio of the sampling factors (\( k_h/k_w \)) to 2:1 to ensure that the sampling areas maintain equidistant shapes and that the extracted patches are proportionally distributed in OI. The number of sampling patches is fixed to 10 for consistency. To analyze the impact of sampling factors on model performance, we evaluate four different sets of sampling factors, as shown in Tab.~\ref{factor}. The results indicate that with an increase in sampling factors, the model's performance improves, suggesting that the size of the sampling region significantly influences the extraction of semantic information. A larger sampling area allows the model to capture richer semantic features, and optimal performance is achieved when the sampling factors are \( \kappa_h = 0.2 \) and \( \kappa_w = 0.1 \). However, further increases in sampling factors lead to a decrease in performance. We suggest that an excessively large sampling area may result in redundancy of semantic information, thereby affecting the model's performance.

\begin{table}[h]
\caption{The performance of VU-BOIQA with different sampling strategies, where P num and V num denote the number of patches and viewports, respectively.}
\setlength{\tabcolsep}{2.4mm}{
\begin{tabular}{@{}llllll@{}}
\toprule
\multirow{2}{*}{Sampling Strategy}   & \multirow{2}{*}{P/V num} & \multicolumn{2}{l}{JUFE-10K} & \multicolumn{2}{l}{OIQ-10K} \\ \cmidrule(l){3-6} 
                                     &                          & PLCC          & SRCC         & PLCC         & SRCC         \\ \cmidrule(r){1-6}
\multirow{2}{*}{Spherical sampling~\cite{fang2022perceptual}}  & 5                        &   0.742            &    0.735          & 0.788             &     0.781         \\
                                     & 10                       &  0.753             &      0.747        & \textbf{0.802}            &\textbf{0.796}              \\
\multirow{2}{*}{Saliency-guided~\cite{xu2020vgcn}}     & 5                        &    0.708           &   0.706           &   0.761           &    0.747          \\
                                     & 10                       &     0.763          &    0.760         &      0.777        &   0.767           \\
\multirow{2}{*}{Equatorial sampling~\cite{yan2024omnidirectional}} & 5        & 0.732              &      0.727         &  0.789            &     0.775         \\
                                     & 10                       &  0.778             &   0.777          & 0.792             &   0.780           \\
\multirow{2}{*}{APS}                 & 5                       &   0.731             & 0.726             &     0.747         &  0.727            \\
                                     & 10                       &  \textbf{0.782}             &    \textbf{0.781}           &     0.766          &   0.749           \\ \cmidrule(l){1-6} 
\end{tabular}}

\label{sampling strategy}
\end{table}

\begin{table}[]
\caption{The influence of loss function.}
\label{loss function}
\setlength{\tabcolsep}{7mm}{
\begin{tabular}{@{}ccccc@{}}
\toprule
\multirow{2}{*}{Loss function} & \multicolumn{2}{c}{JUFE-10K} & \multicolumn{2}{c}{OIQ-10K} \\ \cmidrule(l){2-5} 
  & PLCC    & SRCC  & PLCC & SRCC   \\ \cmidrule(r){1-5}
\textit{L}1  &    \textbf{0.785}   &   \textbf{0.785}  &  0.732   & 0.715      \\
\textit{L}2  &  0.782   &   0.782  &  0.741  & 0.718   \\
norm-in-norm &   0.782  &   0.781  &   \textbf{0.766}  & \textbf{0.749} \\ 
\bottomrule
\end{tabular}}
\label{tab:loss}
\end{table}

\begin{table}[h]
\caption{Performance comparison on two large 2D-IQA databases.}
\setlength{\tabcolsep}{2.3mm}{
\begin{tabular}{@{}ccccccc@{}}
\toprule
\multirow{2}{*}{Method} & \multirow{2}{*}{\begin{tabular}[c]{@{}c@{}}Params\\ (M)\end{tabular}} & \multirow{2}{*}{\begin{tabular}[c]{@{}c@{}}FLOPs\\ (G)\end{tabular}} & \multicolumn{2}{c}{KADID-10K} & \multicolumn{2}{c}{KonIQ-10K} \\ \cmidrule(l){4-7} 
  &   &    & PLCC          & SRCC     & PLCC   & SRCC          \\ \midrule
HyperIQA~\cite{su2020blindly}   & 27.4   & N/A   & 0.872    & 0.869  & 0.900   & \textbf{0.915}         \\
MANIQA~\cite{yang2022maniqa}     & 135.7  & 108.6 & \textbf{0.946}    & \textbf{0.944}  & 0.915   & 0.880         \\
VCRNet~\cite{pan2022vcrnet}     & 16.7   & 10.3  & -   & -    & 0.894    & 0.872         \\
LIQE~\cite{zhang2023blind}       & 151    & 44.2  & 0.930      & 0.931    & \textbf{0.919}    & 0.908    \\\midrule
VU-BOIQA   & 30.2   & 40.8  & 0.824    & 0.807   & 0.848   & 0.797   \\ \bottomrule
\end{tabular}}
\label{iqa compared}
\end{table}

\subsubsection{The effectiveness of each component.}
The experimental results are shown in Tab.~\ref{each component}, where we can clearly observe that a significant performance improvement is achieved by adding the PDFF module, which means that the PDFF module can effectively alleviate the influence of irregular distortion in the ERP image, and its multilevel feature fusion strategy can also promote the interaction of features at different levels and scales. After adding the HPA module, the model performance is further improved. It should be noted that the PA module brings a significant boost on the JUFE-10K database, while almost no performance benefit on the OIQ-10K database, this is mainly attributed to the presence of both uniform and non-uniform distortions in the OIQ-10K database. For OIs with uniform distortion, each patch may contribute similarly, and therefore applying the PA module to allocate different weights to those patches is less valid.

\subsubsection{The influence of sampling strategy.}
We explore the impact of different sampling strategies, including viewport-based spherical sampling method~\cite{fang2022perceptual}, Saliency-guided method~\cite{xu2020vgcn}, Equatorial sampling method~\cite{yan2024subjective} and the APS module, where the number of patches is set to 5 and 10, respectively. The experimental results are shown in Tab.~\ref{sampling strategy}. As a result, spherical sampling, which performs uniform sampling on the sphere, demonstrates superior performance on the OIQ-10K database. In contrast, the proposed APS method focuses on specific regions and may miss the overall distortion features, which can lead to reduced performance. However, APS has the advantage of lower computational complexity and avoids a cumbersome process.

\subsubsection{The comparison of different loss functions.} We evaluate the performance of VU-BOIQA with different loss functions, including $\textit{L}$1, $\textit{L}$2, and norm-in-norm, on the JUFE-10K and OIQ-10K datasets, and the results are shown in Tab.~\ref{tab:loss}. From this table, we can observe that VU-BOIQA behaves stablely and shows better performance with norm-in-norm loss than the other two losses, \ie, $\textit{L}$1 and $\textit{L}$2 on the OIQ-10K database. Thus, we empirically choose the norm-in-norm loss to optimize VU-BOIQA for fast convergence speed and optimization stability.

\subsubsection{The adaptation ability to 2D-IQA} We test the performance of our proposed model on two large-scale 2D image databases, \ie, KADID-10K \cite{lin2019kadid} and KonIQ-10K~\cite{hosu2020koniq}. As shown in Tab.~\ref{iqa compared}, although performing slightly worse than larger models, under similar Params and FLOPs conditions, VU-BOIQA shows a minor performance decrease of 6.3$\%$ on KADID-10K and 9.3$\%$ on KonIQ-10K compared to HyperIQA. In summary, even with relatively few parameters and low computational cost, our model remains capable of handling IQA tasks effectively.

\section{Conclusion and Future Work}
In this paper, we focus on the viewport-reliant problem in BOIQA and propose a new viewport-unaware BOIQA model. Specifically, we design an APS module for extracting representative patches from OIs in ERP format. By doing this, the proposed BOIQA model does not need extract viewports any more. Furthermore, because of the inborn geometry deformation in ERP images, we plug the deformation convolution in multiscale feature extraction to better measure patch-wise quality degradation. Finally, a quality aggregation module is used to map patch-wise quality to global quality. Through extensive experiments, we demonstrate that the proposed model achieves competitive performance and can also be adapted to 2D-IQA. 

Admittedly, there still exists much room for improvement of viewport-unaware BOIQA, where more intelligent deformation processing metrics are desired to be explored in depth. In addition, we plan to bridge the gap between IQA and OIQA, since the proposed VU-BOIQA already shows good applicability for both IQA and OIQA; and we can try to build a more unified and generalizable quality assessment framework for 2D plane images and OIs.
\bibliographystyle{ACM-Reference-Format}
\bibliography{tomm}


\begin{thebibliography}{55}


\ifx \showCODEN    \undefined \def \showCODEN     #1{\unskip}     \fi
\ifx \showDOI      \undefined \def \showDOI       #1{#1}\fi
\ifx \showISBNx    \undefined \def \showISBNx     #1{\unskip}     \fi
\ifx \showISBNxiii \undefined \def \showISBNxiii  #1{\unskip}     \fi
\ifx \showISSN     \undefined \def \showISSN      #1{\unskip}     \fi
\ifx \showLCCN     \undefined \def \showLCCN      #1{\unskip}     \fi
\ifx \shownote     \undefined \def \shownote      #1{#1}          \fi
\ifx \showarticletitle \undefined \def \showarticletitle #1{#1}   \fi
\ifx \showURL      \undefined \def \showURL       {\relax}        \fi
\providecommand\bibfield[2]{#2}
\providecommand\bibinfo[2]{#2}
\providecommand\natexlab[1]{#1}
\providecommand\showeprint[2][]{arXiv:#2}

\bibitem[Bian et~al\mbox{.}(2024)]%
        {bian2024}
\bibfield{author}{\bibinfo{person}{Jiang Bian}, \bibinfo{person}{Xuhong Li}, \bibinfo{person}{Tao Wang}, \bibinfo{person}{Qingzhong Wang}, \bibinfo{person}{Jun Huang}, \bibinfo{person}{Chen Liu}, \bibinfo{person}{Jun Zhao}, \bibinfo{person}{Feixiang Lu}, \bibinfo{person}{Dejing Dou}, {and} \bibinfo{person}{Haoyi Xiong}.} \bibinfo{year}{2024}\natexlab{}.
\newblock \showarticletitle{P2ANet: {A} dataset and benchmark for dense action detection from table tennis match broadcasting videos}.
\newblock \bibinfo{journal}{\emph{ACM Transactions on Multimedia Computing, Communications and Applications}}  \bibinfo{volume}{20} (\bibinfo{year}{2024}), \bibinfo{pages}{23}.
\newblock


\bibitem[Chai et~al\mbox{.}(2021)]%
        {chai2021monocular}
\bibfield{author}{\bibinfo{person}{Xiongli Chai}, \bibinfo{person}{Feng Shao}, \bibinfo{person}{Qiuping Jiang}, \bibinfo{person}{Xiangchao Meng}, {and} \bibinfo{person}{Yo-Sung Ho}.} \bibinfo{year}{2021}\natexlab{}.
\newblock \showarticletitle{Monocular and binocular interactions oriented deformable convolutional networks for blind quality assessment of stereoscopic omnidirectional images}.
\newblock \bibinfo{journal}{\emph{IEEE Transactions on Circuits and Systems for Video Technology}} \bibinfo{volume}{32}, \bibinfo{number}{6} (\bibinfo{year}{2021}), \bibinfo{pages}{3407--3421}.
\newblock


\bibitem[Chen et~al\mbox{.}(2018)]%
        {chen2018spherical}
\bibfield{author}{\bibinfo{person}{Sijia Chen}, \bibinfo{person}{Yingxue Zhang}, \bibinfo{person}{Yiming Li}, \bibinfo{person}{Zhenzhong Chen}, {and} \bibinfo{person}{Zhou Wang}.} \bibinfo{year}{2018}\natexlab{}.
\newblock \showarticletitle{Spherical structural similarity index for objective omnidirectional video quality assessment}. In \bibinfo{booktitle}{\emph{IEEE International Conference on Multimedia and Expo}}. \bibinfo{pages}{1--6}.
\newblock


\bibitem[Dai et~al\mbox{.}(2017)]%
        {dai2017deformable}
\bibfield{author}{\bibinfo{person}{Jifeng Dai}, \bibinfo{person}{Haozhi Qi}, \bibinfo{person}{Yuwen Xiong}, \bibinfo{person}{Yi Li}, \bibinfo{person}{Guodong Zhang}, \bibinfo{person}{Han Hu}, {and} \bibinfo{person}{Yichen Wei}.} \bibinfo{year}{2017}\natexlab{}.
\newblock \showarticletitle{Deformable convolutional networks}. In \bibinfo{booktitle}{\emph{IEEE Conference on Computer Vision and Pattern Recognition}}. \bibinfo{pages}{764--773}.
\newblock


\bibitem[Deng et~al\mbox{.}(2009)]%
        {deng2009imagenet}
\bibfield{author}{\bibinfo{person}{Jia Deng}, \bibinfo{person}{Wei Dong}, \bibinfo{person}{Richard Socher}, \bibinfo{person}{Li-Jia Li}, \bibinfo{person}{Kai Li}, {and} \bibinfo{person}{Li Fei-Fei}.} \bibinfo{year}{2009}\natexlab{}.
\newblock \showarticletitle{ImageNet: {A} large-scale hierarchical image database}. In \bibinfo{booktitle}{\emph{2009 IEEE Conference on Computer Vision and Pattern Recognition}}. \bibinfo{pages}{248--255}.
\newblock


\bibitem[Ding et~al\mbox{.}(2021)]%
        {ding2021comparison}
\bibfield{author}{\bibinfo{person}{Keyan Ding}, \bibinfo{person}{Kede Ma}, \bibinfo{person}{Shiqi Wang}, {and} \bibinfo{person}{Eero~P Simoncelli}.} \bibinfo{year}{2021}\natexlab{}.
\newblock \showarticletitle{Comparison of full-reference image quality models for optimization of image processing systems}.
\newblock \bibinfo{journal}{\emph{International Journal of Computer Vision}} \bibinfo{volume}{129}, \bibinfo{number}{4} (\bibinfo{year}{2021}), \bibinfo{pages}{1258--1281}.
\newblock


\bibitem[Duan et~al\mbox{.}(2018)]%
        {duan2018perceptual}
\bibfield{author}{\bibinfo{person}{Huiyu Duan}, \bibinfo{person}{Guangtao Zhai}, \bibinfo{person}{Xiongkuo Min}, \bibinfo{person}{Yucheng Zhu}, \bibinfo{person}{Yi Fang}, {and} \bibinfo{person}{Xiaokang Yang}.} \bibinfo{year}{2018}\natexlab{}.
\newblock \showarticletitle{Perceptual quality assessment of omnidirectional images}. In \bibinfo{booktitle}{\emph{IEEE International Symposium on Circuits and Systems}}. \bibinfo{pages}{1--5}.
\newblock


\bibitem[Fang et~al\mbox{.}(2022)]%
        {fang2022perceptual}
\bibfield{author}{\bibinfo{person}{Yuming Fang}, \bibinfo{person}{Liping Huang}, \bibinfo{person}{Jiebin Yan}, \bibinfo{person}{Xuelin Liu}, {and} \bibinfo{person}{Yang Liu}.} \bibinfo{year}{2022}\natexlab{}.
\newblock \showarticletitle{Perceptual quality assessment of omnidirectional images}. In \bibinfo{booktitle}{\emph{AAAI Conference on Artificial Intelligence}}, Vol.~\bibinfo{volume}{36}. \bibinfo{pages}{580--588}.
\newblock


\bibitem[Fang et~al\mbox{.}(2017a)]%
        {fang2017no}
\bibfield{author}{\bibinfo{person}{Yuming Fang}, \bibinfo{person}{Jiebin Yan}, \bibinfo{person}{Leida Li}, \bibinfo{person}{Jinjian Wu}, {and} \bibinfo{person}{Weisi Lin}.} \bibinfo{year}{2017}\natexlab{a}.
\newblock \showarticletitle{No reference quality assessment for screen content images with both local and global feature representation}.
\newblock \bibinfo{journal}{\emph{IEEE Transactions on Image Processing}} \bibinfo{volume}{27}, \bibinfo{number}{4} (\bibinfo{year}{2017}), \bibinfo{pages}{1600--1610}.
\newblock


\bibitem[Fang et~al\mbox{.}(2017b)]%
        {fang2017objective}
\bibfield{author}{\bibinfo{person}{Yuming Fang}, \bibinfo{person}{Jiebin Yan}, \bibinfo{person}{Jiaying Liu}, \bibinfo{person}{Shiqi Wang}, \bibinfo{person}{Qiaohong Li}, {and} \bibinfo{person}{Zongming Guo}.} \bibinfo{year}{2017}\natexlab{b}.
\newblock \showarticletitle{Objective quality assessment of screen content images by uncertainty weighting}.
\newblock \bibinfo{journal}{\emph{IEEE Transactions on Image Processing}} \bibinfo{volume}{26}, \bibinfo{number}{4} (\bibinfo{year}{2017}), \bibinfo{pages}{2016--2027}.
\newblock


\bibitem[Fang et~al\mbox{.}(2021)]%
        {fang2021superpixel}
\bibfield{author}{\bibinfo{person}{Yuming Fang}, \bibinfo{person}{Yan Zeng}, \bibinfo{person}{Wenhui Jiang}, \bibinfo{person}{Hanwei Zhu}, {and} \bibinfo{person}{Jiebin Yan}.} \bibinfo{year}{2021}\natexlab{}.
\newblock \showarticletitle{Superpixel-based quality assessment of multi-exposure image fusion for both static and dynamic scenes}.
\newblock \bibinfo{journal}{\emph{IEEE Transactions on Image Processing}}  \bibinfo{volume}{30} (\bibinfo{year}{2021}), \bibinfo{pages}{2526--2537}.
\newblock


\bibitem[Fang et~al\mbox{.}(2020)]%
        {fang2020perceptual}
\bibfield{author}{\bibinfo{person}{Yuming Fang}, \bibinfo{person}{Hanwei Zhu}, \bibinfo{person}{Yan Zeng}, \bibinfo{person}{Kede Ma}, {and} \bibinfo{person}{Zhou Wang}.} \bibinfo{year}{2020}\natexlab{}.
\newblock \showarticletitle{Perceptual quality assessment of smartphone photography}. In \bibinfo{booktitle}{\emph{IEEE Conference on Computer Vision and Pattern Recognition}}. \bibinfo{pages}{3677--3686}.
\newblock


\bibitem[Hosu et~al\mbox{.}(2020)]%
        {hosu2020koniq}
\bibfield{author}{\bibinfo{person}{Vlad Hosu}, \bibinfo{person}{Hanhe Lin}, \bibinfo{person}{Tamas Sziranyi}, {and} \bibinfo{person}{Dietmar Saupe}.} \bibinfo{year}{2020}\natexlab{}.
\newblock \showarticletitle{KonIQ-10k: {An} ecologically valid database for deep learning of blind image quality assessment}.
\newblock \bibinfo{journal}{\emph{IEEE Transactions on Image Processing}}  \bibinfo{volume}{29} (\bibinfo{year}{2020}), \bibinfo{pages}{4041--4056}.
\newblock


\bibitem[Jiang et~al\mbox{.}(2021)]%
        {jiang2021cubemap}
\bibfield{author}{\bibinfo{person}{Hao Jiang}, \bibinfo{person}{Gangyi Jiang}, \bibinfo{person}{Mei Yu}, \bibinfo{person}{Yun Zhang}, \bibinfo{person}{You Yang}, \bibinfo{person}{Zongju Peng}, \bibinfo{person}{Fen Chen}, {and} \bibinfo{person}{Qingbo Zhang}.} \bibinfo{year}{2021}\natexlab{}.
\newblock \showarticletitle{Cubemap-based perception-driven blind quality assessment for 360-degree images}.
\newblock \bibinfo{journal}{\emph{IEEE Transactions on Image Processing}}  \bibinfo{volume}{30} (\bibinfo{year}{2021}), \bibinfo{pages}{2364--2377}.
\newblock


\bibitem[Kim et~al\mbox{.}(2019)]%
        {kim2019deep}
\bibfield{author}{\bibinfo{person}{Hak~Gu Kim}, \bibinfo{person}{Heoun-Taek Lim}, {and} \bibinfo{person}{Yong~Man Ro}.} \bibinfo{year}{2019}\natexlab{}.
\newblock \showarticletitle{Deep virtual reality image quality assessment with human perception guider for omnidirectional image}.
\newblock \bibinfo{journal}{\emph{IEEE Transactions on Circuits and Systems for Video Technology}} \bibinfo{volume}{30}, \bibinfo{number}{4} (\bibinfo{year}{2019}), \bibinfo{pages}{917--928}.
\newblock


\bibitem[Li et~al\mbox{.}(2019)]%
        {li2019viewport}
\bibfield{author}{\bibinfo{person}{Chen Li}, \bibinfo{person}{Mai Xu}, \bibinfo{person}{Lai Jiang}, \bibinfo{person}{Shanyi Zhang}, {and} \bibinfo{person}{Xiaoming Tao}.} \bibinfo{year}{2019}\natexlab{}.
\newblock \showarticletitle{Viewport proposal CNN for 360° video quality assessment}. In \bibinfo{booktitle}{\emph{IEEE Conference on Computer Vision and Pattern Recognition}}. \bibinfo{pages}{10169--10178}.
\newblock


\bibitem[Li et~al\mbox{.}(2020)]%
        {li2020norm}
\bibfield{author}{\bibinfo{person}{Dingquan Li}, \bibinfo{person}{Tingting Jiang}, {and} \bibinfo{person}{Ming Jiang}.} \bibinfo{year}{2020}\natexlab{}.
\newblock \showarticletitle{Norm-in-norm loss with faster convergence and better performance for image quality assessment}. In \bibinfo{booktitle}{\emph{ACM International Conference on Multimedia}}. \bibinfo{pages}{789--797}.
\newblock


\bibitem[Li and Zhang(2023)]%
        {li2023mfan}
\bibfield{author}{\bibinfo{person}{Huanyang Li} {and} \bibinfo{person}{Xinfeng Zhang}.} \bibinfo{year}{2023}\natexlab{}.
\newblock \showarticletitle{{MFAN}: {A} Multi-projection fusion attention network for no-reference and full-reference panoramic image quality assessment}.
\newblock \bibinfo{journal}{\emph{IEEE Signal Processing Letters}}  \bibinfo{volume}{30} (\bibinfo{year}{2023}), \bibinfo{pages}{1207--1211}.
\newblock


\bibitem[Li et~al\mbox{.}(2021)]%
        {li2021pseudocylindrical}
\bibfield{author}{\bibinfo{person}{Mu Li}, \bibinfo{person}{Kede Ma}, \bibinfo{person}{Jinxing Li}, {and} \bibinfo{person}{David Zhang}.} \bibinfo{year}{2021}\natexlab{}.
\newblock \showarticletitle{Pseudocylindrical convolutions for learned omnidirectional image compression}.
\newblock \bibinfo{journal}{\emph{arXiv preprint arXiv:2112.13227}} (\bibinfo{year}{2021}).
\newblock


\bibitem[Lim et~al\mbox{.}(2018)]%
        {lim2018vr}
\bibfield{author}{\bibinfo{person}{Heaun-Taek Lim}, \bibinfo{person}{Hak~Gu Kim}, {and} \bibinfo{person}{Yang~Man Ra}.} \bibinfo{year}{2018}\natexlab{}.
\newblock \showarticletitle{VR IQA NET: {D}eep virtual reality image quality assessment using adversarial learning}. In \bibinfo{booktitle}{\emph{IEEE International Conference on Acoustics, Speech and Signal Processing}}. \bibinfo{pages}{6737--6741}.
\newblock


\bibitem[Lin et~al\mbox{.}(2019)]%
        {lin2019kadid}
\bibfield{author}{\bibinfo{person}{Hanhe Lin}, \bibinfo{person}{Vlad Hosu}, {and} \bibinfo{person}{Dietmar Saupe}.} \bibinfo{year}{2019}\natexlab{}.
\newblock \showarticletitle{KADID-10k: {A} large-scale artificially distorted IQA database}. In \bibinfo{booktitle}{\emph{2019 Eleventh International Conference on Quality of Multimedia Experience}}. \bibinfo{pages}{1--3}.
\newblock


\bibitem[Liu et~al\mbox{.}(2024a)]%
        {liu2024perceptual}
\bibfield{author}{\bibinfo{person}{Xuelin Liu}, \bibinfo{person}{Jiebin Yan}, \bibinfo{person}{Liping Huang}, \bibinfo{person}{Yuming Fang}, \bibinfo{person}{Zheng Wan}, {and} \bibinfo{person}{Yang Liu}.} \bibinfo{year}{2024}\natexlab{a}.
\newblock \showarticletitle{Perceptual quality assessment of omnidirectional images: {A} benchmark and computational model}.
\newblock \bibinfo{journal}{\emph{ACM Transactions on Multimedia Computing, Communications and Applications}} \bibinfo{volume}{20}, \bibinfo{number}{6} (\bibinfo{year}{2024}), \bibinfo{pages}{1--24}.
\newblock


\bibitem[Liu et~al\mbox{.}(2024b)]%
        {liu2024transdiff}
\bibfield{author}{\bibinfo{person}{Xiaoxiao Liu}, \bibinfo{person}{Yan Zhao}, \bibinfo{person}{Shigang Wang}, {and} \bibinfo{person}{Jian Wei}.} \bibinfo{year}{2024}\natexlab{b}.
\newblock \showarticletitle{TransDiff: {M}edical image segmentation method based on Swin Transformer with diffusion probabilistic model}.
\newblock \bibinfo{journal}{\emph{Applied Intelligence}} \bibinfo{volume}{54}, \bibinfo{number}{8} (\bibinfo{year}{2024}), \bibinfo{pages}{6543--6557}.
\newblock


\bibitem[Liu et~al\mbox{.}(2023)]%
        {liu2023toward}
\bibfield{author}{\bibinfo{person}{Yun Liu}, \bibinfo{person}{Xiaohua Yin}, \bibinfo{person}{Zuliang Wan}, \bibinfo{person}{Guanghui Yue}, {and} \bibinfo{person}{Zhi Zheng}.} \bibinfo{year}{2023}\natexlab{}.
\newblock \showarticletitle{Toward a no-reference omnidirectional image quality evaluation by using multi-perceptual features}.
\newblock \bibinfo{journal}{\emph{ACM Transactions on Multimedia Computing, Communications and Applications}} \bibinfo{volume}{19}, \bibinfo{number}{2} (\bibinfo{year}{2023}), \bibinfo{pages}{1--19}.
\newblock


\bibitem[Liu et~al\mbox{.}(2022)]%
        {liu2022swin}
\bibfield{author}{\bibinfo{person}{Ze Liu}, \bibinfo{person}{Han Hu}, \bibinfo{person}{Yutong Lin}, \bibinfo{person}{Zhuliang Yao}, \bibinfo{person}{Zhenda Xie}, \bibinfo{person}{Yixuan Wei}, \bibinfo{person}{Jia Ning}, \bibinfo{person}{Yue Cao}, \bibinfo{person}{Zheng Zhang}, \bibinfo{person}{Li Dong}, {et~al\mbox{.}}} \bibinfo{year}{2022}\natexlab{}.
\newblock \showarticletitle{Swin transformer v2: {Scaling} up capacity and resolution}. In \bibinfo{booktitle}{\emph{IEEE Conference on Computer Vision and Pattern Recognition}}. \bibinfo{pages}{12009--12019}.
\newblock


\bibitem[Pan et~al\mbox{.}(2022)]%
        {pan2022vcrnet}
\bibfield{author}{\bibinfo{person}{Zhaoqing Pan}, \bibinfo{person}{Feng Yuan}, \bibinfo{person}{Jianjun Lei}, \bibinfo{person}{Yuming Fang}, \bibinfo{person}{Xiao Shao}, {and} \bibinfo{person}{Sam Kwong}.} \bibinfo{year}{2022}\natexlab{}.
\newblock \showarticletitle{{VCRNet}: {Visual} compensation restoration network for no-reference image quality assessment}.
\newblock \bibinfo{journal}{\emph{IEEE Transactions on Image Processing}}  \bibinfo{volume}{31} (\bibinfo{year}{2022}), \bibinfo{pages}{1613--1627}.
\newblock


\bibitem[Su et~al\mbox{.}(2020)]%
        {su2020blindly}
\bibfield{author}{\bibinfo{person}{Shaolin Su}, \bibinfo{person}{Qingsen Yan}, \bibinfo{person}{Yu Zhu}, \bibinfo{person}{Cheng Zhang}, \bibinfo{person}{Xin Ge}, \bibinfo{person}{Jinqiu Sun}, {and} \bibinfo{person}{Yanning Zhang}.} \bibinfo{year}{2020}\natexlab{}.
\newblock \showarticletitle{Blindly assess image quality in the wild guided by a self-adaptive hyper network}. In \bibinfo{booktitle}{\emph{IEEE Conference on Computer Vision and Pattern Recognition}}. \bibinfo{pages}{3667--3676}.
\newblock


\bibitem[Sui et~al\mbox{.}(2023)]%
        {sui2023scandmm}
\bibfield{author}{\bibinfo{person}{Xiangjie Sui}, \bibinfo{person}{Yuming Fang}, \bibinfo{person}{Hanwei Zhu}, \bibinfo{person}{Shiqi Wang}, {and} \bibinfo{person}{Zhou Wang}.} \bibinfo{year}{2023}\natexlab{}.
\newblock \showarticletitle{ScanDMM: {A} deep markov model of scanpath prediction for 360deg images}. In \bibinfo{booktitle}{\emph{IEEE Conference on Computer Vision and Pattern Recognition}}. \bibinfo{pages}{6989--6999}.
\newblock


\bibitem[Sui et~al\mbox{.}(2021)]%
        {sui2021perceptual}
\bibfield{author}{\bibinfo{person}{Xiangjie Sui}, \bibinfo{person}{Kede Ma}, \bibinfo{person}{Yiru Yao}, {and} \bibinfo{person}{Yuming Fang}.} \bibinfo{year}{2021}\natexlab{}.
\newblock \showarticletitle{Perceptual quality assessment of omnidirectional images as moving camera videos}.
\newblock \bibinfo{journal}{\emph{IEEE Transactions on Visualization and Computer Graphics}} \bibinfo{volume}{28}, \bibinfo{number}{8} (\bibinfo{year}{2021}), \bibinfo{pages}{3022--3034}.
\newblock


\bibitem[Sun et~al\mbox{.}(2018)]%
        {sun2018large}
\bibfield{author}{\bibinfo{person}{Wei Sun}, \bibinfo{person}{Ke Gu}, \bibinfo{person}{Siwei Ma}, \bibinfo{person}{Wenhan Zhu}, \bibinfo{person}{Ning Liu}, {and} \bibinfo{person}{Guangtao Zhai}.} \bibinfo{year}{2018}\natexlab{}.
\newblock \showarticletitle{A large-scale compressed 360-degree spherical image database: {F}rom subjective quality evaluation to objective model comparison}. In \bibinfo{booktitle}{\emph{IEEE 20th International Workshop on Multimedia Signal Processing}}. \bibinfo{pages}{1--6}.
\newblock


\bibitem[Sun et~al\mbox{.}(2019)]%
        {sun2019mc360iqa}
\bibfield{author}{\bibinfo{person}{Wei Sun}, \bibinfo{person}{Xiongkuo Min}, \bibinfo{person}{Guangtao Zhai}, \bibinfo{person}{Ke Gu}, \bibinfo{person}{Huiyu Duan}, {and} \bibinfo{person}{Siwei Ma}.} \bibinfo{year}{2019}\natexlab{}.
\newblock \showarticletitle{MC360IQA: {A} multi-channel CNN for blind 360-degree image quality assessment}.
\newblock \bibinfo{journal}{\emph{IEEE Journal of Selected Topics in Signal Processing}} \bibinfo{volume}{14}, \bibinfo{number}{1} (\bibinfo{year}{2019}), \bibinfo{pages}{64--77}.
\newblock


\bibitem[Sun et~al\mbox{.}(2017)]%
        {sun2017weighted}
\bibfield{author}{\bibinfo{person}{Yule Sun}, \bibinfo{person}{Ang Lu}, {and} \bibinfo{person}{Lu Yu}.} \bibinfo{year}{2017}\natexlab{}.
\newblock \showarticletitle{Weighted-to-spherically-uniform quality evaluation for omnidirectional video}.
\newblock \bibinfo{journal}{\emph{IEEE Signal Processing Letters}} \bibinfo{volume}{24}, \bibinfo{number}{9} (\bibinfo{year}{2017}), \bibinfo{pages}{1408--1412}.
\newblock


\bibitem[VQEG(2000)]%
        {vqeg}
\bibfield{author}{\bibinfo{person}{VQEG}.} \bibinfo{year}{2000}\natexlab{}.
\newblock \bibinfo{title}{Final report from the video quality experts group on the validation of objective models of video quality assessment}.
\newblock \bibinfo{howpublished}{\url{http://www.vqeg.org}}.
\newblock
\newblock
\shownote{Accessed: 2021-06-17}.


\bibitem[Wang et~al\mbox{.}(2023)]%
        {wang2023internimage}
\bibfield{author}{\bibinfo{person}{Wenhai Wang}, \bibinfo{person}{Jifeng Dai}, \bibinfo{person}{Zhe Chen}, \bibinfo{person}{Zhenhang Huang}, \bibinfo{person}{Zhiqi Li}, \bibinfo{person}{Xizhou Zhu}, \bibinfo{person}{Xiaowei Hu}, \bibinfo{person}{Tong Lu}, \bibinfo{person}{Lewei Lu}, \bibinfo{person}{Hongsheng Li}, {et~al\mbox{.}}} \bibinfo{year}{2023}\natexlab{}.
\newblock \showarticletitle{Internimage: {Exploring} large-scale vision foundation models with deformable convolutions}. In \bibinfo{booktitle}{\emph{IEEE Conference on Computer Vision and Pattern Recognition}}. \bibinfo{pages}{14408--14419}.
\newblock


\bibitem[Wang et~al\mbox{.}(2004)]%
        {wang2004image}
\bibfield{author}{\bibinfo{person}{Zhou Wang}, \bibinfo{person}{Alan~C Bovik}, \bibinfo{person}{Hamid~R Sheikh}, {and} \bibinfo{person}{Eero~P Simoncelli}.} \bibinfo{year}{2004}\natexlab{}.
\newblock \showarticletitle{Image quality assessment: {F}rom error visibility to structural similarity}.
\newblock \bibinfo{journal}{\emph{IEEE Transactions on Image Processing}} \bibinfo{volume}{13}, \bibinfo{number}{4} (\bibinfo{year}{2004}), \bibinfo{pages}{600--612}.
\newblock


\bibitem[Wang and Rehman(2017)]%
        {wang2017begin}
\bibfield{author}{\bibinfo{person}{Zhou Wang} {and} \bibinfo{person}{Abdul Rehman}.} \bibinfo{year}{2017}\natexlab{}.
\newblock \showarticletitle{Begin with the end in mind: {A} unified end-to-end quality-of-experience monitoring, optimization and management framework}. In \bibinfo{booktitle}{\emph{SMPTE Annual Technical Conference and Exhibition}}. \bibinfo{pages}{1--11}.
\newblock


\bibitem[Woo et~al\mbox{.}(2018)]%
        {woo2018cbam}
\bibfield{author}{\bibinfo{person}{Sanghyun Woo}, \bibinfo{person}{Jongchan Park}, \bibinfo{person}{Joon-Young Lee}, {and} \bibinfo{person}{In~So Kweon}.} \bibinfo{year}{2018}\natexlab{}.
\newblock \showarticletitle{CBAM: {C}onvolutional block attention module}. In \bibinfo{booktitle}{\emph{European Conference on Computer Vision}}. \bibinfo{pages}{3--19}.
\newblock


\bibitem[Wu et~al\mbox{.}(2024)]%
        {wu2024assessor360}
\bibfield{author}{\bibinfo{person}{Tianhe Wu}, \bibinfo{person}{Shuwei Shi}, \bibinfo{person}{Haoming Cai}, \bibinfo{person}{Mingdeng Cao}, \bibinfo{person}{Jing Xiao}, \bibinfo{person}{Yinqiang Zheng}, {and} \bibinfo{person}{Yujiu Yang}.} \bibinfo{year}{2024}\natexlab{}.
\newblock \showarticletitle{Assessor360: {M}ulti-sequence network for blind omnidirectional image quality assessment}. In \bibinfo{booktitle}{\emph{Advances in Neural Information Processing Systems}}. \bibinfo{pages}{64957--64970}.
\newblock


\bibitem[Xu et~al\mbox{.}(2019)]%
        {xu2019quality}
\bibfield{author}{\bibinfo{person}{Jiahua Xu}, \bibinfo{person}{Ziyuan Luo}, \bibinfo{person}{Wei Zhou}, \bibinfo{person}{Wenyuan Zhang}, {and} \bibinfo{person}{Zhibo Chen}.} \bibinfo{year}{2019}\natexlab{}.
\newblock \showarticletitle{Quality assessment of stereoscopic 360-degree images from multi-viewports}. In \bibinfo{booktitle}{\emph{Picture Coding Symposium}}. \bibinfo{pages}{1--5}.
\newblock


\bibitem[Xu et~al\mbox{.}(2020)]%
        {xu2020vgcn}
\bibfield{author}{\bibinfo{person}{Jiahua Xu}, \bibinfo{person}{Wei Zhou}, {and} \bibinfo{person}{Zhibo Chen}.} \bibinfo{year}{2020}\natexlab{}.
\newblock \showarticletitle{Blind omnidirectional image quality assessment with viewport oriented graph convolutional networks}.
\newblock \bibinfo{journal}{\emph{IEEE Transactions on Circuits and Systems for Video Technology}} \bibinfo{volume}{31}, \bibinfo{number}{5} (\bibinfo{year}{2020}), \bibinfo{pages}{1724--1737}.
\newblock


\bibitem[Yan et~al\mbox{.}(2020)]%
        {yan2020no}
\bibfield{author}{\bibinfo{person}{Jiebin Yan}, \bibinfo{person}{Yuming Fang}, \bibinfo{person}{Rengang Du}, \bibinfo{person}{Yan Zeng}, {and} \bibinfo{person}{Yifan Zuo}.} \bibinfo{year}{2020}\natexlab{}.
\newblock \showarticletitle{No reference quality assessment for {3D} synthesized views by local structure variation and global naturalness change}.
\newblock \bibinfo{journal}{\emph{IEEE Transactions on Image Processing}}  \bibinfo{volume}{29} (\bibinfo{year}{2020}), \bibinfo{pages}{7443--7453}.
\newblock


\bibitem[Yan et~al\mbox{.}(2022)]%
        {yan2022subjective}
\bibfield{author}{\bibinfo{person}{Jiebin Yan}, \bibinfo{person}{Jing Li}, \bibinfo{person}{Yuming Fang}, \bibinfo{person}{Zhaohui Che}, \bibinfo{person}{Xue Xia}, {and} \bibinfo{person}{Yang Liu}.} \bibinfo{year}{2022}\natexlab{}.
\newblock \showarticletitle{Subjective and objective quality of experience of free viewpoint videos}.
\newblock \bibinfo{journal}{\emph{IEEE Transactions on Image Processing}}  \bibinfo{volume}{31} (\bibinfo{year}{2022}), \bibinfo{pages}{3896--3907}.
\newblock


\bibitem[Yan et~al\mbox{.}(pted)]%
        {yan2024subjective}
\bibfield{author}{\bibinfo{person}{Jiebin Yan}, \bibinfo{person}{Jiale Rao}, \bibinfo{person}{Xuelin Liu}, \bibinfo{person}{Yuming Fang}, \bibinfo{person}{Yifan Zuo}, {and} \bibinfo{person}{Weide Liu}.} \bibinfo{year}{2024, accepted}\natexlab{}.
\newblock \showarticletitle{Subjective and objective quality assessment of non-uniformly distorted omnidirectional images}.
\newblock \bibinfo{journal}{\emph{IEEE Transactions on Multimedia}} (\bibinfo{year}{2024, accepted}).
\newblock


\bibitem[Yan et~al\mbox{.}(2024)]%
        {yan2024omnidirectional}
\bibfield{author}{\bibinfo{person}{Jiebin Yan}, \bibinfo{person}{Ziwen Tan}, \bibinfo{person}{Yuming Fang}, \bibinfo{person}{Junjie Chen}, \bibinfo{person}{Wenhui Jiang}, {and} \bibinfo{person}{Zhou Wang}.} \bibinfo{year}{2024}\natexlab{}.
\newblock \showarticletitle{Omnidirectional image quality captioning: {A} large-scale database and a new model}.
\newblock \bibinfo{journal}{\emph{IEEE Transactions on Image Processing}}  \bibinfo{volume}{34} (\bibinfo{year}{2024}), \bibinfo{pages}{1326--1339}.
\newblock


\bibitem[Yang et~al\mbox{.}(2022)]%
        {yang2022maniqa}
\bibfield{author}{\bibinfo{person}{Sidi Yang}, \bibinfo{person}{Tianhe Wu}, \bibinfo{person}{Shuwei Shi}, \bibinfo{person}{Shanshan Lao}, \bibinfo{person}{Yuan Gong}, \bibinfo{person}{Mingdeng Cao}, \bibinfo{person}{Jiahao Wang}, {and} \bibinfo{person}{Yujiu Yang}.} \bibinfo{year}{2022}\natexlab{}.
\newblock \showarticletitle{{MANIQA}: {Multi-dimension} attention network for no-reference image quality assessment}. In \bibinfo{booktitle}{\emph{IEEE Conference on Computer Vision and Pattern Recognition}}. \bibinfo{pages}{1191--1200}.
\newblock


\bibitem[Ye et~al\mbox{.}(2024)]%
        {ye2024segment}
\bibfield{author}{\bibinfo{person}{Jiabo Ye}, \bibinfo{person}{Junfeng Tian}, \bibinfo{person}{Ming Yan}, \bibinfo{person}{Haiyang Xu}, \bibinfo{person}{Qinghao Ye}, \bibinfo{person}{Yaya Shi}, \bibinfo{person}{Xiaoshan Yang}, \bibinfo{person}{Xuwu Wang}, \bibinfo{person}{Ji Zhang}, \bibinfo{person}{Liang He}, {and} \bibinfo{person}{Xin Lin}.} \bibinfo{year}{2024}\natexlab{}.
\newblock \showarticletitle{UniQRNet: {U}nifying referring expression grounding and segmentation with QRNet}.
\newblock \bibinfo{journal}{\emph{ACM Transactions on Multimedia Computing, Communications and Applications}}  \bibinfo{volume}{20} (\bibinfo{year}{2024}), \bibinfo{pages}{28}.
\newblock


\bibitem[Yu et~al\mbox{.}(2015)]%
        {yu2015framework}
\bibfield{author}{\bibinfo{person}{Matt Yu}, \bibinfo{person}{Haricharan Lakshman}, {and} \bibinfo{person}{Bernd Girod}.} \bibinfo{year}{2015}\natexlab{}.
\newblock \showarticletitle{A framework to evaluate omnidirectional video coding schemes}. In \bibinfo{booktitle}{\emph{IEEE International Symposium on Mixed and Augmented Reality}}. \bibinfo{pages}{31--36}.
\newblock


\bibitem[Zakharchenko et~al\mbox{.}(2016)]%
        {zakharchenko2016quality}
\bibfield{author}{\bibinfo{person}{Vladyslav Zakharchenko}, \bibinfo{person}{Kwang~Pyo Choi}, {and} \bibinfo{person}{Jeong~Hoon Park}.} \bibinfo{year}{2016}\natexlab{}.
\newblock \showarticletitle{Quality metric for spherical panoramic video}. In \bibinfo{booktitle}{\emph{Optics and Photonics for Information Processing X}}, Vol.~\bibinfo{volume}{9970}. \bibinfo{pages}{57--65}.
\newblock


\bibitem[Zhang and Liu(2022)]%
        {zhang2022no}
\bibfield{author}{\bibinfo{person}{Chaofan Zhang} {and} \bibinfo{person}{Shiguang Liu}.} \bibinfo{year}{2022}\natexlab{}.
\newblock \showarticletitle{No-reference omnidirectional image quality assessment based on joint network}. In \bibinfo{booktitle}{\emph{ACM International Conference on Multimedia}}. \bibinfo{pages}{943--951}.
\newblock


\bibitem[Zhang et~al\mbox{.}(2023)]%
        {zhang2023blind}
\bibfield{author}{\bibinfo{person}{Weixia Zhang}, \bibinfo{person}{Guangtao Zhai}, \bibinfo{person}{Ying Wei}, \bibinfo{person}{Xiaokang Yang}, {and} \bibinfo{person}{Kede Ma}.} \bibinfo{year}{2023}\natexlab{}.
\newblock \showarticletitle{Blind image quality assessment via vision-language correspondence: {A} multitask learning perspective}. In \bibinfo{booktitle}{\emph{IEEE Conference on Computer Vision and Pattern Recognition}}. \bibinfo{pages}{14071--14081}.
\newblock


\bibitem[Zheng et~al\mbox{.}(2020)]%
        {zheng2020segmented}
\bibfield{author}{\bibinfo{person}{Xuelei Zheng}, \bibinfo{person}{Gangyi Jiang}, \bibinfo{person}{Mei Yu}, {and} \bibinfo{person}{Hao Jiang}.} \bibinfo{year}{2020}\natexlab{}.
\newblock \showarticletitle{Segmented spherical projection-based blind omnidirectional image quality assessment}.
\newblock \bibinfo{journal}{\emph{IEEE Access}}  \bibinfo{volume}{8} (\bibinfo{year}{2020}), \bibinfo{pages}{31647--31659}.
\newblock


\bibitem[Zhou et~al\mbox{.}(2023)]%
        {zhou360degree}
\bibfield{author}{\bibinfo{person}{Mingliang Zhou}, \bibinfo{person}{Lei Chen}, \bibinfo{person}{Xuekai Wei}, \bibinfo{person}{Xingran Liao}, \bibinfo{person}{Qin Mao}, \bibinfo{person}{Heqiang Wang}, \bibinfo{person}{Huayan Pu}, \bibinfo{person}{Jun Luo}, \bibinfo{person}{Tao Xiang}, {and} \bibinfo{person}{Bin Fang}.} \bibinfo{year}{2023}\natexlab{}.
\newblock \showarticletitle{Perception-oriented u-shaped transformer network for 360-degree no-reference image quality assessment}.
\newblock \bibinfo{journal}{\emph{IEEE Transactions on Broadcasting}}  \bibinfo{volume}{69} (\bibinfo{year}{2023}), \bibinfo{pages}{396--450}.
\newblock


\bibitem[Zhou et~al\mbox{.}(2021)]%
        {zhou2021omnidirectional}
\bibfield{author}{\bibinfo{person}{Yu Zhou}, \bibinfo{person}{Yanjing Sun}, \bibinfo{person}{Leida Li}, \bibinfo{person}{Ke Gu}, {and} \bibinfo{person}{Yuming Fang}.} \bibinfo{year}{2021}\natexlab{}.
\newblock \showarticletitle{Omnidirectional image quality assessment by distortion discrimination assisted multi-stream network}.
\newblock \bibinfo{journal}{\emph{IEEE Transactions on Circuits and Systems for Video Technology}} \bibinfo{volume}{32}, \bibinfo{number}{4} (\bibinfo{year}{2021}), \bibinfo{pages}{1767--1777}.
\newblock


\bibitem[Zhou et~al\mbox{.}(2018)]%
        {zhou2018weighted}
\bibfield{author}{\bibinfo{person}{Yufeng Zhou}, \bibinfo{person}{Mei Yu}, \bibinfo{person}{Hualin Ma}, \bibinfo{person}{Hua Shao}, {and} \bibinfo{person}{Gangyi Jiang}.} \bibinfo{year}{2018}\natexlab{}.
\newblock \showarticletitle{Weighted-to-spherically-uniform SSIM objective quality evaluation for panoramic video}. In \bibinfo{booktitle}{\emph{IEEE International Conference on Signal Processing}}. \bibinfo{pages}{54--57}.
\newblock


\bibitem[Zhu et~al\mbox{.}(2019)]%
        {zhu2019deformable}
\bibfield{author}{\bibinfo{person}{Xizhou Zhu}, \bibinfo{person}{Han Hu}, \bibinfo{person}{Stephen Lin}, {and} \bibinfo{person}{Jifeng Dai}.} \bibinfo{year}{2019}\natexlab{}.
\newblock \showarticletitle{Deformable convnets v2: {M}ore deformable, better results}. In \bibinfo{booktitle}{\emph{IEEE Conference on Computer Vision and Pattern Recognition}}. \bibinfo{pages}{9308--9316}.
\newblock


\end{thebibliography}


\end{document}